\documentclass[10pt]{article}
\PassOptionsToPackage{hyphens}{url}   

\usepackage{amsmath}
\usepackage{scicom}
\usepackage[authoryear]{natbib}

\usepackage{hyperref}
\usepackage{bookmark}
\usepackage{url}
\usepackage{nicefrac}
\usepackage{microtype}
\usepackage{tabularx}
\usepackage{multirow}
\usepackage{float}
\usepackage{adjustbox}
\usepackage{longtable}
\usepackage{array}
\usepackage[section]{placeins}

\newcolumntype{L}[1]{>{\raggedright\arraybackslash}p{#1}}
\input{numbers}

\hypersetup{
pdftitle={Where Words Begin and End},
pdfauthor={Husein Zolkepli},
pdfsubject={Audio and Speech Processing},
pdfkeywords={forced alignment, word timestamps, CTC, telephone speech, codecs, multilingual, Malay},
colorlinks=true,
linkcolor=scLink,
citecolor=scLink,
urlcolor=scLink
}
\title{Where Words Begin and End}
\scshorttitle{Where Words Begin and End}
\scauthor{Husein Zolkepli}{husein.zolkepli@scicom.com.my}
\scaffiliation{Scicom (MSC) Berhad, Malaysia}
\scdate{October 2026}

\begin{document}

\maketitle

\begin{abstract}
Forced aligners locate each word of a known transcript in the audio, and many speech pipelines depend on them. In our
own data, a widely used CTC aligner inserts a pause at nearly every word boundary, and we found almost no aligner
evaluated on telephone audio. We compare nine aligner configurations against human word boundaries on read,
conversational and telephone English. The Montreal Forced Aligner (MFA) is the most precise, while CTC aligners
consistently start words late, because most frames inside a word go to the blank token. We then fine-tune Whisper with
LoRA to predict each word's start and end as timestamp tokens, training only on boundaries placed by people:
\val{fa/train-hours} hours in \val{fa/train-languages} languages. The resulting aligner is more precise than MFA on
every English test condition, and than MFA's own Korean and Dutch models. With FA-Bench's protocol and code on TIMIT,
its word-boundary error is \val{fb/ts/mae} ms, against 18.2 ms for the best published system. Two choices
matter most on telephone conversations: placing the timestamp slots after the word they time, and training on noisy
transcripts, which also makes the aligner nearly insensitive to errors in the transcript. The released model reaches
\val{swb/ft-ts-v3-final/tel/mae} ms on real landline calls, against \val{s2/bar/swb/mae} ms for MFA. The models are
available at \url{https://huggingface.co/collections/Scicom-intl/force-alignment}.
\end{abstract}

\section{Introduction}

A forced aligner takes a recording and its transcript and returns where each word starts and ends. Text-to-speech
corpora are cut at these boundaries, end-of-turn detectors learn from the pauses between them, and subtitles,
pronunciation scoring, data filtering and interleaved speech-text pretraining take them as given. In our own stack
one aligner supplies all of them: MMS-300M \citep{pratap2024scaling} with CTC decoding, run through the defaults of the
\texttt{ctc-forced-aligner} package. It sits behind our transcription service's word timestamps, the Malaysian speech
corpora built with it, our end-of-turn labels and the quality gate of our speech synthesis data.

Those defaults are wrong in a way that propagates. Across \val{prior/words-m} million words of the published
Malaysian corpora aligned with them, between \val{prior/gap-any-min}\% and \val{prior/gap-any-max}\% of word
boundaries carry a gap, and \val{prior/gap01-min}\% to \val{prior/gap01-max}\% carry one of at least 0.1 s, a pause
by any end-of-turn labeller's standard. Read speech aligned with the Montreal Forced Aligner (MFA) has a gap at
\val{prior/mfa-gap-any-min}\% to \val{prior/mfa-gap-any-max}\% of its boundaries. Checked against a voice activity
detector, \val{prior/mid-speech-min}\% to \val{prior/mid-speech-max}\% of the Malaysian pauses have their midpoint
inside speech. The aligner invents a pause between almost every pair of words, and its words are shorter than they
sound.

We want a replacement that is precise against human boundaries rather than another aligner, covers the languages of
a Malaysian call centre (Malay, English, Mandarin, Tamil, Cantonese, Indonesian) and the switching between them
without a pronunciation dictionary per language, and holds up on telephone speech: 8 kHz, codecs, line noise. This
report asks what exists, how precise it is and why, and then trains a replacement on boundaries that people placed.

It makes five contributions. First, a way to measure alignment precision on telephone speech that keeps the reference
time base: real codec binaries, with each codec's delay measured and removed (Section~\ref{sec:channel}). MP3 at 8 kHz
alone arrives \val{timit/lag/mp3} ms late, more than the mean error of any pilot aligner on clean read speech (at most
\val{timit/clean/mae-max} ms). Second, a pilot benchmark of nine aligner configurations at their defaults against human
word boundaries: TIMIT and Buckeye in clean and simulated telephone conditions, and Switchboard conversations recorded
over real landlines (Section~\ref{sec:pilot}). Third, a measurement of why CTC aligners misplace boundaries, taken on
the forced path itself (Section~\ref{sec:ctc}). Fourth, a survey of aligners and of evaluation data with human
boundaries, which locates the gaps (Sections~\ref{sec:aligners} and~\ref{sec:data}). Fifth, aligners trained only on
human word boundaries in \val{fa/train-languages} languages (Sections~\ref{sec:design} and~\ref{sec:results}).
Whisper-Scicom predicting word timestamps beats MFA on every set where MFA has a model, and most of that skill
transfers from English alone. The timestamp slots must come after the word they time, telephone augmentation carries
over to codecs never seen in training, and neither the adapter configuration nor a finer timestamp vocabulary matters
(Section~\ref{sec:conclusion} sums up what moved the error). The models trained on the public Whisper checkpoints,
large-v3 and large-v3-turbo, are released with a standalone aligner script\footnote{\url{https://huggingface.co/Scicom-intl/Where-Words-Begin-and-End-Whisper-Large-V3} and
\url{https://huggingface.co/Scicom-intl/Where-Words-Begin-and-End-Whisper-Large-V3-Turbo}, CC BY-NC 4.0.}.

\section{Background}
\label{sec:background}

\subsection{How aligners place words}

\textbf{Hidden Markov models.} The classic aligner trains Gaussian-mixture HMMs over a pronunciation dictionary and
Viterbi-decodes the known word sequence: P2FA, Prosodylab-Aligner, EasyAlign and MAUS on HTK, Gentle and MFA on Kaldi
\citep{yuan2008speaker,gorman2011prosodylabaligner,goldman2011easyalign,kisler2017multilingual,ochshorn2017gentle,mcauliffe2017montreal}.
Every frame belongs to some phone, so a boundary is a change of state on a 10 ms grid, and a three-state phone lasts
at least 30 ms \citep{rousso2024tradition}. Independent comparisons keep putting MFA ahead of the neural aligners
built on speech recognisers at tight tolerances
\citep{rousso2024tradition,mcauliffe2026montreal,chu2026fabench,zeyer2026gradientbased}. The price is a dictionary and
an acoustic model per language: the MFA model listing has none for Malay or Indonesian
\citep{montrealcorpustools2024mfa}, and run per language over FLEURS it fails on about a fifth of the utterances
\citep{akavarapu2026phoneme}. Gentle is the one widely used open aligner whose acoustic model is natively 8 kHz
\citep{ochshorn2017gentle}.

\textbf{CTC.} CTC-segmentation aligns text to frame-wise CTC posteriors \citep{kurzinger2020ctcsegmentation}. MMS used
the same idea to segment Bible recordings in 1{,}130 languages, with text romanised by uroman and a \texttt{<star>}
token, scored with probability one, to absorb audio the text does not cover (chapter introductions, digits)
\citep{pratap2024scaling}. Its alignments were judged only through downstream error rates, never against boundaries.
torchaudio ships the model and a GPU Viterbi \citep{hwang2023torchaudio}; \texttt{ctc-forced-aligner} wraps the same
checkpoint and inserts the star before every word by default \citep{ashraf2024ctcforcedaligner}. The NeMo Forced
Aligner runs Viterbi on CTC log-probabilities of NeMo models \citep{rastorgueva2023nemo}, and WhisperX aligns Whisper's
transcript with a wav2vec2 character CTC model \citep{bain2023whisperx}.

\textbf{Attention.} Whisper's word timestamps come from dynamic time warping over the cross-attention of a few
alignment heads, on its 20 ms encoder grid \citep{radford2023robust}; stable-ts and whisper-timestamped wrap the same
mechanism \citep{jian2022stablets,louradour2023whispertimestamped}. Evaluations of these timestamps have mostly used
collars of 200 ms or more \citep{yeh2025whisper,zusag2024crisperwhisper}.

\textbf{Learned boundaries.} A third line trains on boundaries directly. Charsiu classifies 10 ms frames into phones
trained on MFA alignments, because wav2vec2's 20 ms frames are too coarse for precise timing \citep{zhu2022phonetoaudio};
NeuFA, Wav2TextGrid, MAPS and FALCON learn from human boundaries \citep{li2022neufa,kadambi2025tunable,kelley2024masonalberta,rousso2026fully};
FuseAlign refines word edges on a 1 ms grid, trained on 78k hours of Gentle labels with low-pass augmentation
\citep{vaidya2026fusealign}. The newest are speech LLMs that fill timestamp slots: Qwen3-ForcedAligner-0.6B predicts
indices on an 80 ms grid, trained on MFA labels, in 11 languages
\citep{shi2026qwen3asr,mu2026llmforcedaligner}, and Granite-Speech predicts 10 ms frame indices as digit tokens
\citep{kons2026word}. The common recipe in 2025 and 2026 is to distil an HMM teacher into a neural student
\citep{shi2026qwen3asr,kons2026word,wagner2026transcription,vaidya2026fusealign}.

\subsection{How precision is measured}

Papers disagree on what they score: starts, ends or both, tolerance accuracy at 10 to 100 ms, mean absolute error
under three definitions of accumulated average shift, or precision and recall inside a collar
\citep{shi2023achieving,rousso2024tradition,chu2026fabench}. We report starts and ends separately and both pooled.
What counts as precise is set by people. Four Buckeye transcribers agreed exactly on 43\% of word boundaries and
placed them 26 ms apart on average; on phones, 79\% of their boundaries fell within 20 ms of each other
\citep{raymond2002analysis}. Three labellers of spontaneous German agreed on 93\% of phone boundaries within 20 ms
\citep{kipp1997pronunciation}. On in-the-wild English, pairs of annotators placed only 64\% of word boundaries within
25 ms of each other \citep{vaidya2026fusealign}. On telephone speech the only figure we found is from the OGI corpus: labellers agreed
within 10 ms on 78\% of the boundaries both put in the same broad class \citep{cole1994labeler}. An aligner at 80 to
90\% of boundaries within 20 to 25 ms on clean speech is at human agreement; beyond that a benchmark measures its
annotators.

\textbf{Telephone speech is unmeasured.} We found no aligner, of any family, evaluated against human boundaries on
8 kHz or codec-processed speech. FA-Bench, the broadest benchmark so far, adds noise, music, babble and reverberation
\citep{chu2026fabench}; WhisperX scores Switchboard, but on its own transcripts with a 200 ms collar
\citep{bain2023whisperx}; the one telephone voice in a 1997 study of synthesis-based alignment kept 48.6\% of
boundaries within 20 ms \citep{malfrere1997highquality}.

\subsection{Why CTC aligners misplace boundaries}
\label{sec:ctc}

A CTC model emits one of its units or a blank at every frame, and training asks only that the units come out in the
right order, not where the sound is. The models learn to fire each character on one frame or
two and to give every other frame to the blank \citep{graves2006connectionist,zeyer2021why}. A forced aligner then decodes the
known character sequence through those emissions, and a word's start and end are read off its first and last
character frame.

\begin{figure}[H]
\centering
\includegraphics[width=0.92\linewidth]{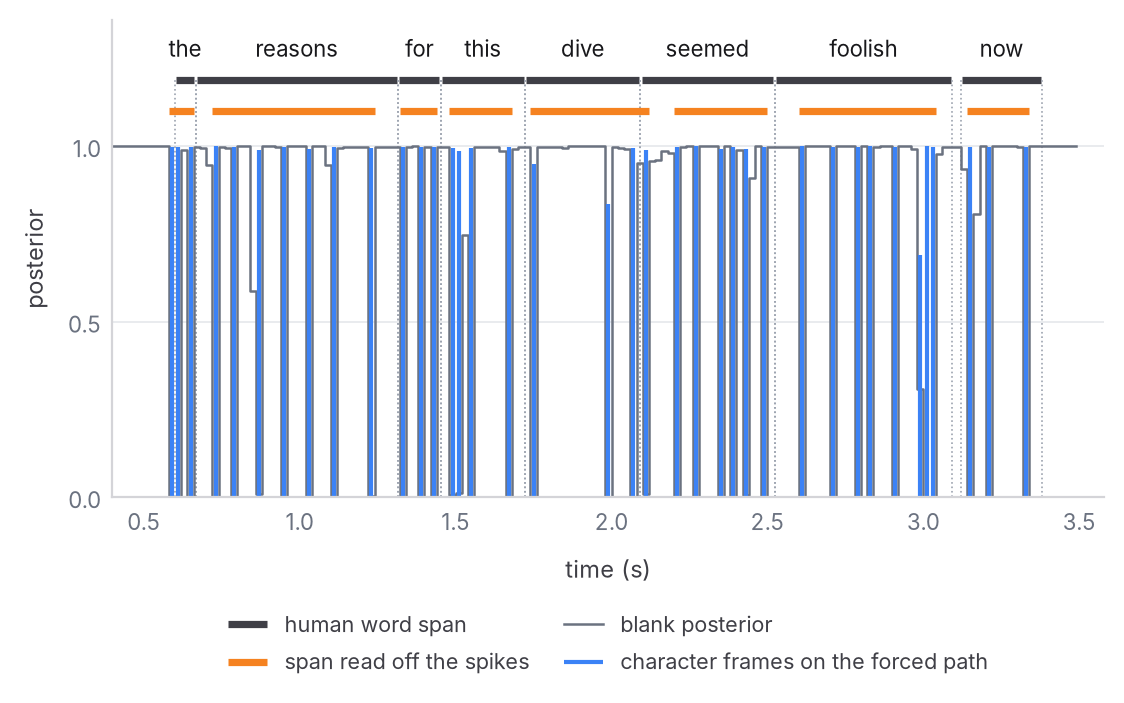}
\caption{MMS-300M emissions over one clean TIMIT test sentence (\texttt{DR1\_AKS0\_SI2203}). Dark bars: the human word
spans from TIMIT's phone labels. Orange bars: each word's span as a decoder reads it off the forced path, from the
first character frame to the end of the last. Grey: the blank posterior. Blue: the frames the forced path gives to a
character.}
\label{fig:ctcpeaky}
\end{figure}

Figure~\ref{fig:ctcpeaky} shows the result on one clean TIMIT sentence with MMS-300M. The blank posterior sits near
one almost everywhere, inside words too, and the character frames on the forced path are thin spikes whose span is
shorter than the word a human marked, with gaps where the speech is continuous.

\begin{figure}[H]
\centering
\includegraphics[width=0.92\linewidth]{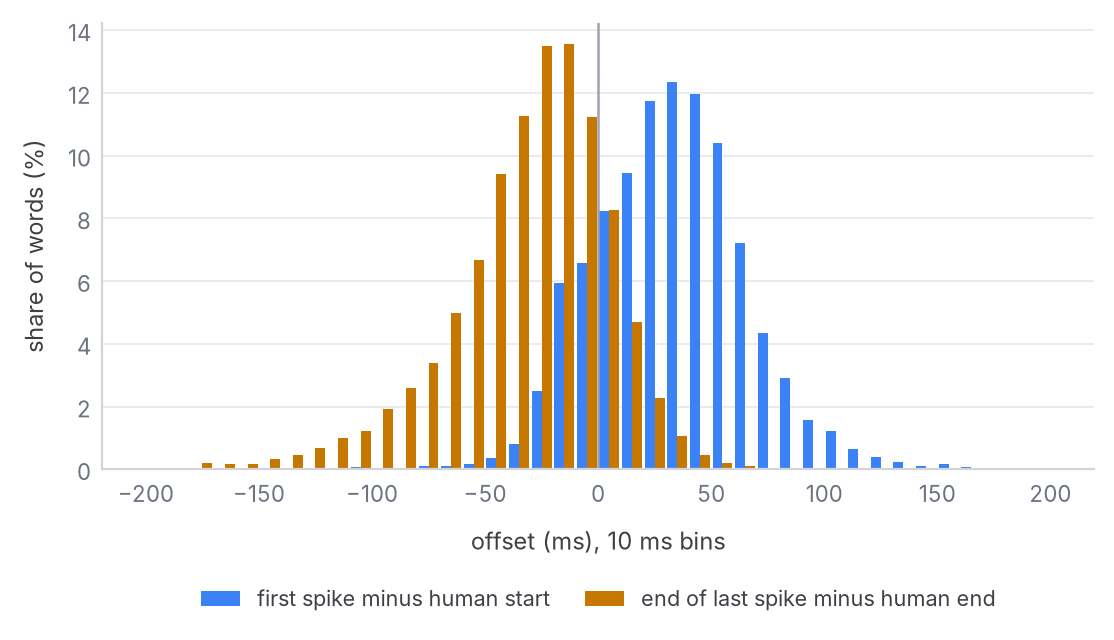}
\caption{Where a word's first and last character frame fall relative to its human start and end, MMS-300M forced paths
on clean TIMIT test, 10 ms bins.}
\label{fig:ctcedges}
\end{figure}

Over all \val{peaky/clean/n-words} words of the clean test set, a median \val{peaky/clean/blank}\% of the frames inside
a human word are blank on the forced path. The first character frame falls a median \val{peaky/clean/first} ms after
the human start (interquartile range \val{peaky/clean/first-q25} to \val{peaky/clean/first-q75} ms), and the last
character frame ends a median \val{peaky/clean/last} ms before the human end (Figure~\ref{fig:ctcedges}). Any decoder
that reads edges off the spikes inherits these offsets, and decoding tricks move them around without removing them:
the model never learned where the word was. Nor do they depend on the channel. Under G.711 the first frame still falls
a median \val{peaky/g711u/first} ms after the human start, and under GSM \val{peaky/gsm/first} ms.

The two settings of \texttt{ctc-forced-aligner} that our corpora used make this worse in opposite directions. A
\texttt{<star>} wildcard before every word, scored with probability one on every frame, takes every frame it can, so
each word keeps only the frames between its first and last spike. The package then pads each word with half of the
neighbouring blank and the last word with all of it. Our transcription service ports a different decoder, the
torchaudio tutorial's trellis, with the same star column; both are in the pilot (Section~\ref{sec:pilot}).

\subsection{What a boundary is}
\label{sec:boundary}

A boundary error is only as meaningful as the boundary. TIMIT's word boundaries come from its hand-labelled phone
boundaries, and TIMIT counts the closure of a stop, the silent interval before the burst, as part of the stop. A word
that starts with a stop therefore starts at the onset of silence, where nothing in the signal marks it. If late starts
concentrated on those words, part of an aligner's error would be the convention, but they do not. With MMS-300M, words
that start with a vowel start a median \val{class/mms-torchaudio/onset/vowel} ms late and words that start with a
fricative \val{class/mms-torchaudio/onset/fricative} ms late; words that start with a closure start
\val{class/mms-torchaudio/onset/closure} ms late, and words that start on a burst, a sharp event in the signal, only
\val{class/mms-torchaudio/onset/burst} ms late. The convention adds to the error at closures; the spike timing makes
most of it. MFA, whose stop models include the closure, starts the same closure-initial words a median
\val{class/mfa-arpa/onset/closure} ms late.

\section{Available aligners}
\label{sec:aligners}

We surveyed \val{survey/n-models} systems, open and commercial; Table~\ref{tab:models} lists the open aligners that
matter here. Breadth is not what is missing: the MMS models cover every target language through romanisation, and so
do the tools built on Whisper. Precision in those languages is: no system we found, open or commercial, publishes the
timing accuracy of its word boundaries in Malay, Indonesian or Tamil. The aligners that are precise
in English either need a dictionary and an acoustic model per language, and MFA has neither for Malay or Indonesian,
or stop at a fixed set of languages, eleven for Qwen3-ForcedAligner, none of them Malay, Indonesian or Tamil.

\begin{table}[H]
\centering
\caption{Open aligners that matter for this work, from our survey (72 systems reviewed). \emph{Grid} is the resolution of the output timestamps. The language columns mark whether the system returns timestamps for Malay, Indonesian, English, Mandarin, Cantonese and Tamil ($\bullet$ yes, $\cdot$ no, ? not documented). \emph{Lexicon} marks systems that need a pronunciation dictionary. None reports an evaluation against human boundaries on telephone speech.}
\label{tab:models}
\footnotesize
\begin{adjustbox}{max width=\linewidth}
\begin{tabular}{llrrccccccc}
\toprule
\schead \scth{aligner} & \scth{family} & \scth{grid, ms} & \scth{lang.} & \scth{ms} & \scth{id} & \scth{en} & \scth{zh} & \scth{yue} & \scth{ta} & \scth{lexicon} \\
\midrule
Qwen3-ForcedAligner-0.6B & NAR LLM & 80 & 11 & $\cdot$ & $\cdot$ & $\bullet$ & $\bullet$ & $\bullet$ & $\cdot$ & no \\
FunASR fa-zh & NAR LLM & -- & 1 & $\cdot$ & $\cdot$ & $\cdot$ & $\bullet$ & $\cdot$ & $\cdot$ & no \\
torchaudio \texttt{MMS\_FA} & CTC & 20 & 1{,}130 & $\bullet$ & $\bullet$ & $\bullet$ & $\bullet$ & $\bullet$ & $\bullet$ & no \\
\texttt{ctc-forced-aligner} & CTC & 20 & 1{,}130 & $\bullet$ & $\bullet$ & $\bullet$ & $\bullet$ & $\bullet$ & $\bullet$ & no \\
MMS-1B-all & CTC & 20 & 1{,}162 & $\bullet$ & $\bullet$ & $\bullet$ & $\bullet$ & $\bullet$ & $\bullet$ & no \\
Omnilingual ASR CTC & CTC & 20 & 1{,}600 & $\bullet$ & $\bullet$ & $\bullet$ & $\bullet$ & $\bullet$ & $\bullet$ & no \\
NeMo Forced Aligner & CTC & -- & -- & ? & ? & $\bullet$ & ? & ? & ? & no \\
WhisperX & CTC & 20 & 40 & $\cdot$ & $\bullet$ & $\bullet$ & $\bullet$ & $\cdot$ & $\cdot$ & no \\
stable-ts & attention & 20 & 99 & $\bullet$ & $\bullet$ & $\bullet$ & $\bullet$ & $\bullet$ & $\bullet$ & no \\
MFA 3.x & HMM & 10 & 42 & $\cdot$ & $\cdot$ & $\bullet$ & $\bullet$ & $\cdot$ & $\bullet$ & yes \\
Charsiu & frames & 10 & 2 & $\cdot$ & $\cdot$ & $\bullet$ & $\bullet$ & $\cdot$ & $\cdot$ & no \\
Malaya-Speech aligners & CTC & 20 & 2 & $\bullet$ & ? & $\bullet$ & $\cdot$ & $\cdot$ & $\cdot$ & no \\
\bottomrule
\end{tabular}
\end{adjustbox}
\end{table}

\begin{figure}[H]
\centering
\includegraphics[width=0.9\linewidth]{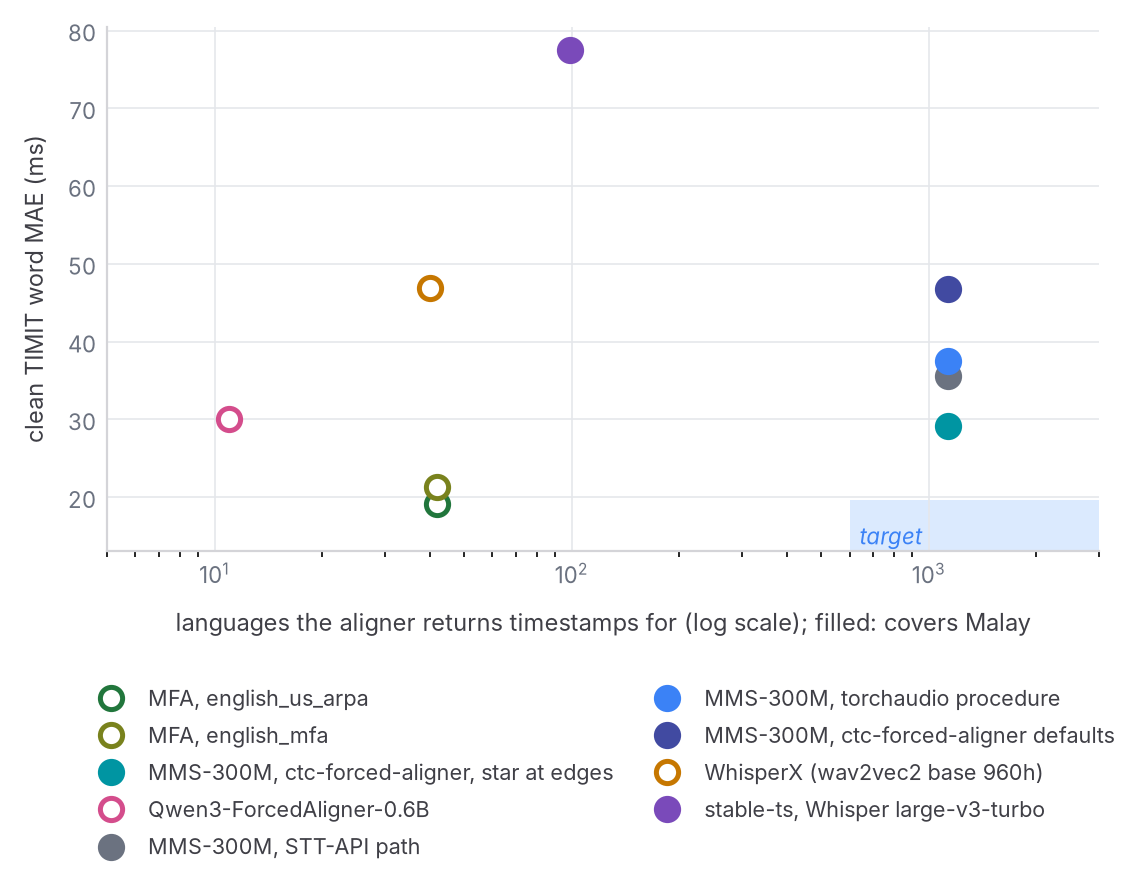}
\caption{Precision against breadth. Vertical: mean absolute word-boundary error on clean TIMIT test from our pilot
(Section~\ref{sec:pilot}). Horizontal: languages for which the aligner returns timestamps, from our survey. Filled
markers cover Malay. The MMS decoders share one model and so one horizontal position.}
\label{fig:landscape}
\end{figure}

Figure~\ref{fig:landscape} puts the aligners we ran on one plane: how many languages each returns timestamps for,
against how far its word boundaries fall from TIMIT's human ones. The corner we want, the breadth of the CTC models at
the precision of the best HMM aligner, is empty.

\section{Evaluation data}
\label{sec:data}

Precision can only be claimed against human boundaries, and those are scarce (Table~\ref{tab:evalsets}). English has
TIMIT and Buckeye; AlignBench, a hand-labelled in-the-wild set, has been described but not released
\citep{vaidya2026fusealign}. Mandarin and Tamil have small or read sets behind licence agreements:
300 hand-labelled utterances of Mandarin broadcast news in MCPST, read words and sentences in SSNCE. Cantonese has 126
isolated words in VoxAngeles. Malay and Indonesian have nothing we could find. The test sets behind Qwen3-ForcedAligner's
multilingual numbers are labelled by MFA, apart from one internal Chinese set, and are not released.

\begin{table}[H]
\centering
\caption{Evaluation data with human or human-corrected time boundaries, from our survey (38 sets reviewed). \emph{Corrected}: automatic boundaries corrected by hand; \emph{TIMIT's}: the telephone versions of TIMIT carry TIMIT's own labels; \emph{manual, 3x}: three annotators. We found no set for Malay or Indonesian.}
\label{tab:evalsets}
\footnotesize
\begin{adjustbox}{max width=\linewidth}
\begin{tabular}{lllllrl}
\toprule
\schead \scth{set} & \scth{languages} & \scth{speech} & \scth{channel} & \scth{boundaries} & \scth{size} & \scth{access} \\
\midrule
TIMIT & en & read & wideband & corrected & 5.4 h & LDC93S1 \\
Buckeye & en & conversation & wideband & corrected & 38.1 h & registration \\
AlignBench & en & in the wild & wideband & manual, 3x & 0.83 h & not released \\
\midrule
NTIMIT & en & read & landline & TIMIT's & 5 h & LDC93S2 \\
STC-TIMIT & en & read & phone line & TIMIT's & 6{,}300 utt. & LDC2008S03 \\
CTIMIT & en & read & cellular & TIMIT's & 3{,}367 utt. & LDC96S30 \\
WTIMIT & en & read & mobile & TIMIT's & 6{,}300 utt. & LDC2010S02 \\
Switchboard ICSI & en & conversation & landline & manual & 3.75 h & ISIP mirror \\
OGI-MLTS & 11 (zh, ta) & monologue & landline & corrected & 38.5 h & LDC94S17 \\
\midrule
MCPST & zh & broadcast & wideband & manual (300) & 7{,}849 utt. & LDC2015S05 \\
Global TIMIT & zh & read & wideband & forced & 5 h & LDC2021S03 \\
SSNCE & ta & read & wideband & corrected & 8 h & LDC2021S04 \\
VoxAngeles & 95 (yue) & words & field & corrected & 1.5 h & CC BY-NC \\
DoReCo 2.0 & 53 & narratives & field & corrected & 100 h & per dataset \\
\bottomrule
\end{tabular}
\end{adjustbox}
\end{table}

Telephone speech with human boundaries exists in English: TIMIT replayed through landline, cellular and mobile
networks, which keep TIMIT's labels and need their own registration checked before any tolerance below 20 ms is
trusted, and the Switchboard Transcription Project, in which transcribers time-marked the phones, syllables and words
of real landline conversations. Of these only the last can be downloaded without a licence agreement, from a public
mirror, and we use its first phase in the pilot. OGI-MLTS adds real telephone speech in eleven languages, Mandarin and
Tamil among them, but its corrected boundaries are mostly broad phonetic classes.

The Singapore National Speech Corpus has telephone recordings in our region, but its
time marks stop at the utterance \citep{koh2019nsc}. Training material is not the constraint: transcribed speech in every target language runs
to thousands of hours, but its word times come from aligners, and a student trained on them can at best match its
teacher.

\section{Telephone channel}
\label{sec:channel}

A codec that delays the signal by 20 ms moves every reference boundary by 20 ms, and the aligner takes the blame. We
therefore build every telephone condition from the
clean recording in five steps: a zero-phase 300 to 3{,}400 Hz band-pass (no group delay), a linear-phase resample to
8 kHz, a round trip through a real codec binary, a cross-correlation against the band-limited reference that measures
the codec's delay and removes it, and a resample back to 16 kHz, the rate every aligner takes
(Table~\ref{tab:channel}).

\begin{table}[H]
\centering
\caption{The telephone conditions. Every one starts from the same zero-phase 300 to 3{,}400 Hz band-pass and an 8 kHz resample, then round-trips the codec binary named. \emph{Delay} is the codec's lag measured against the band-limited reference by cross-correlation, the same on all 1{,}344 utterances; \emph{residual} is the largest lag left after compensation.}
\label{tab:channel}
\footnotesize
\begin{adjustbox}{max width=\linewidth}
\begin{tabular}{lrlrr}
\toprule
\schead \scth{condition} & \scth{kbit/s} & \scth{tool} & \scth{delay, ms} & \scth{residual, ms} \\
\midrule
band-limited only & 128 & -- & 0.000 & 0.0 \\
G.711 $\mu$-law & 64 & ffmpeg & 0.000 & 0.0 \\
G.726 ADPCM & 32 & ffmpeg & 0.000 & 0.0 \\
GSM 06.10 full rate & 13 & sox & 0.000 & 0.0 \\
Opus (SILK, VoIP mode) & 12 & ffmpeg & 0.125 & 0.0 \\
MP3 (LAME) & 16 & ffmpeg & 138.125 & 0.0 \\
\bottomrule
\end{tabular}
\end{adjustbox}
\end{table}

Most codecs add no delay that survives decoding. MP3 does: at 8 kHz the LAME encoder's 576 samples and the decoder's
529 arrive as one constant lag of \val{timit/lag/mp3} ms on every utterance, which ffmpeg does not trim for this
sample rate. Call recordings are commonly archived as low-rate MP3, so the same shift sits in any label derived from
such an archive unless someone removes it. After compensation the residual lag is zero on every utterance.

\section{Pilot benchmark}
\label{sec:pilot}

\subsection{Setup}

\textbf{Data.} TIMIT test without the SA sentences: \val{timit/n-utt} read utterances, \val{timit/n-words} words with
human boundaries, \val{timit/hours} hours, from the Hugging Face mirror \texttt{nh0znoisung/timit}. Buckeye
\citep{pitt2005buckeye}, conversational interviews: the \texttt{alexwengg/buckeye} benchmark carries the corpus's
hand-corrected word times, but its segments run up to 25 s and keep untranscribed interviewer turns between words, so
we cut them again at every pause of at least 0.3 s and draw 50 sub-segments per speaker: \val{buckeye/n-utt}
utterances, \val{buckeye/n-words} words, \val{buckeye/hours} hours. Switchboard, real landline conversations: the
first phase of the ICSI transcription project, in which transcribers time-marked every phone, keeping only utterances
whose every segment is one plain word (no merged words, fragments or noises): \val{swb/n-utt} utterances,
\val{swb/n-words} words, \val{swb/hours} hours of native 8 kHz audio. TIMIT runs clean and in the six conditions of
Table~\ref{tab:channel}, Buckeye clean and in G.711, GSM and Opus, Switchboard as recorded.

\textbf{Aligners.} Each runs at the defaults its documentation shows, because defaults are what people run, on a
shared GPU server (MFA on its CPUs). MFA 3.4.2
\citep{mcauliffe2017montreal} with its \texttt{english\_us\_arpa} and \texttt{english\_mfa} models. Qwen3-ForcedAligner-0.6B
\citep{shi2026qwen3asr}. WhisperX \citep{bain2023whisperx} with its default English model. stable-ts on Whisper
large-v3-turbo \citep{radford2023robust}, which aligns by dynamic time warping over cross-attention. And four decoders
of the same MMS-300M forced-alignment model: the torchaudio procedure (no star token), \texttt{ctc-forced-aligner} at
its defaults (a star before every word), the same with a star only at the edges, and the decoder our transcription
service runs, ported line for line.

\textbf{Scoring.} Every word contributes its start and its end. We report the mean absolute error over all boundaries
(MAE, the quantity Qwen3-ASR calls accumulated average shift), the share of boundaries within 20, 50 and 100 ms, and
the median signed error of starts and of ends. Each condition is scored on the utterances every aligner aligned.

\subsection{Clean speech}

Table~\ref{tab:pilot} ranks the aligners on clean TIMIT. MFA with its \texttt{english\_us\_arpa} model is the most
precise: \val{timit/mfa-arpa/clean/w20}\% of word boundaries within 20 ms and a mean error of
\val{timit/mfa-arpa/clean/mae} ms, in line with what its authors and FA-Bench report on the same corpus
\citep{mcauliffe2026montreal,chu2026fabench}. The best multilingual aligners sit near 30 ms: Qwen3-ForcedAligner at
\val{timit/qwen3fa/clean/mae} ms and the MMS model decoded with the star only at the edges at
\val{timit/cfa-edges/clean/mae} ms. Qwen3-ForcedAligner places only \val{timit/qwen3fa/clean/w20}\% of boundaries
within 20 ms against MFA's \val{timit/mfa-arpa/clean/w20}\%; on an 80 ms grid, closer is out of reach. Whisper's
attention, through stable-ts, is the least precise of all at \val{timit/stablets/clean/mae} ms, with
\val{timit/stablets/clean/w20}\% of boundaries within 20 ms.

\begin{table}[H]
\centering
\caption{Existing aligners on clean TIMIT test (SI and SX sentences, human word boundaries), each at its defaults. Scored on the 1{,}344 utterances every aligner aligned. \emph{MAE} is the mean absolute error over word starts and ends; the tolerance columns give the share of boundaries within 20, 50 and 100 ms; \emph{bias} is the median signed error of starts and of ends (positive is late). \emph{Fail} counts utterances the aligner returned no usable alignment for; \emph{RTF} is compute time over audio time on one shared GPU (MFA: CPU, 16 jobs). \emph{MFA, arpa} and \emph{MFA, mfa}: MFA 3.4 with \texttt{english\_us\_arpa} and \texttt{english\_mfa}; \emph{MMS}: the MMS-300M forced-alignment model decoded by the torchaudio procedure, by \texttt{ctc-forced-aligner} (CFA) at its defaults or with the star only at the edges, or by our transcription service (STT-API); \emph{Qwen3-FA}: Qwen3-ForcedAligner-0.6B; \emph{stable-ts} on Whisper large-v3-turbo.}
\label{tab:pilot}
\footnotesize
\begin{adjustbox}{max width=\linewidth}
\begin{tabular}{lrrrrrrrr}
\toprule
\schead & & \multicolumn{3}{c}{\scth{within, \%}} & \multicolumn{2}{c}{\scth{bias, ms}} & & \\
\cmidrule(lr){3-5}\cmidrule(lr){6-7}
\schead \scth{aligner} & \scth{MAE, ms} & \scth{20 ms} & \scth{50 ms} & \scth{100 ms} & \scth{start} & \scth{end} & \scth{fail} & \scth{RTF} \\
\midrule
MFA, arpa & \textbf{19.1} & \textbf{68.2} & \textbf{91.9} & \textbf{98.3} & $-3$ & $+2$ & 0 & 0.021 \\
MFA, mfa & 21.3 & 64.5 & 90.2 & 97.9 & $-3$ & $+2$ & 0 & 0.011 \\
MMS, CFA edges & 29.2 & 45.5 & 85.9 & 97.1 & $+10$ & $+10$ & 0 & 0.007 \\
Qwen3-FA & 30.1 & 41.0 & 84.6 & 97.5 & $-4$ & $-2$ & 0 & 0.028 \\
MMS, STT-API & 35.6 & 35.4 & 76.4 & 96.2 & $+30$ & $-18$ & 0 & 0.044 \\
MMS, torchaudio & 37.5 & 35.0 & 71.4 & 96.1 & $+40$ & $-18$ & 0 & 0.007 \\
MMS, CFA default & 46.8 & 31.0 & 68.5 & 91.5 & $+34$ & $-22$ & 0 & 0.007 \\
WhisperX & 46.9 & 29.5 & 60.7 & 91.1 & $+60$ & $+1$ & 0 & 0.033 \\
stable-ts & 77.5 & 20.6 & 47.3 & 74.0 & $-22$ & $-51$ & 0 & 0.041 \\
\bottomrule
\end{tabular}
\end{adjustbox}
\end{table}

\begin{figure}[H]
\centering
\includegraphics[width=0.9\linewidth]{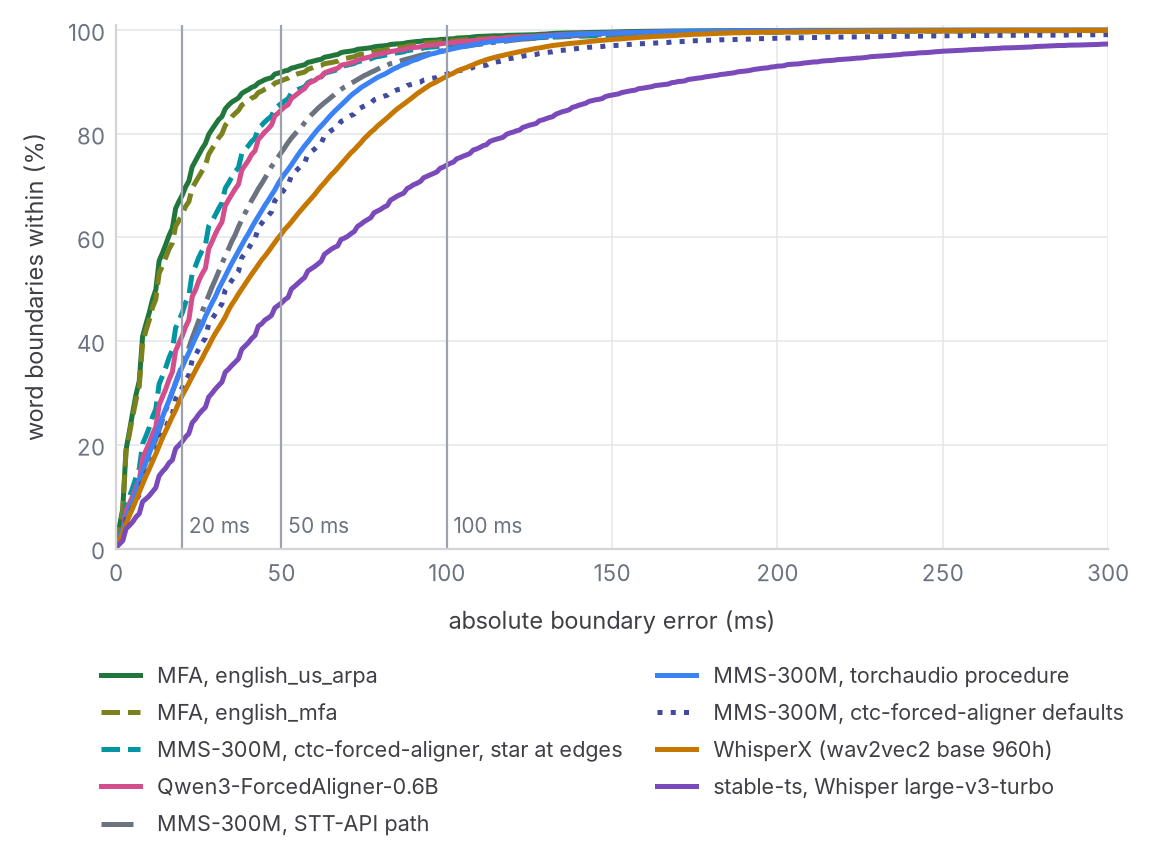}
\caption{Share of TIMIT word boundaries within a given error, clean test set, every aligner at its defaults.}
\label{fig:pilotcdf}
\end{figure}

The four MMS rows read the same emissions of the same model, yet their errors run from
\val{timit/cfa-edges/clean/mae} to \val{timit/cfa-segment/clean/mae} ms: the decoder decides as much as the model.
The package default, a star before every word, is the worst of the four, and our transcription service's decoder
scores \val{timit/sttapi/clean/mae} ms. The difference sits at the edges of the utterance. With a star before every word, the
last word ends a median \val{edges/timit/cfa-segment/last-end/median} ms after the human end, against a mean error of
\val{edges/timit/cfa-segment/interior/mean} ms inside the utterance: the package pads the last word with the whole
trailing blank, the failure that ran Malaysian words to the end of the recording. With the star only at the edges,
the same package is the most precise MMS decoder inside the utterance, \val{edges/timit/cfa-edges/interior/mean} ms
against Qwen3-ForcedAligner's \val{edges/timit/qwen3fa/interior/mean}, but its errors still gather at the edges: on
Switchboard, where clips are short and carry silence at both ends, its last word ends a median
\val{edges/swb/cfa-edges/last-end/median} ms late. Figure~\ref{fig:pilotcdf} shows the whole error distribution.

\begin{figure}[H]
\centering
\includegraphics[width=0.92\linewidth]{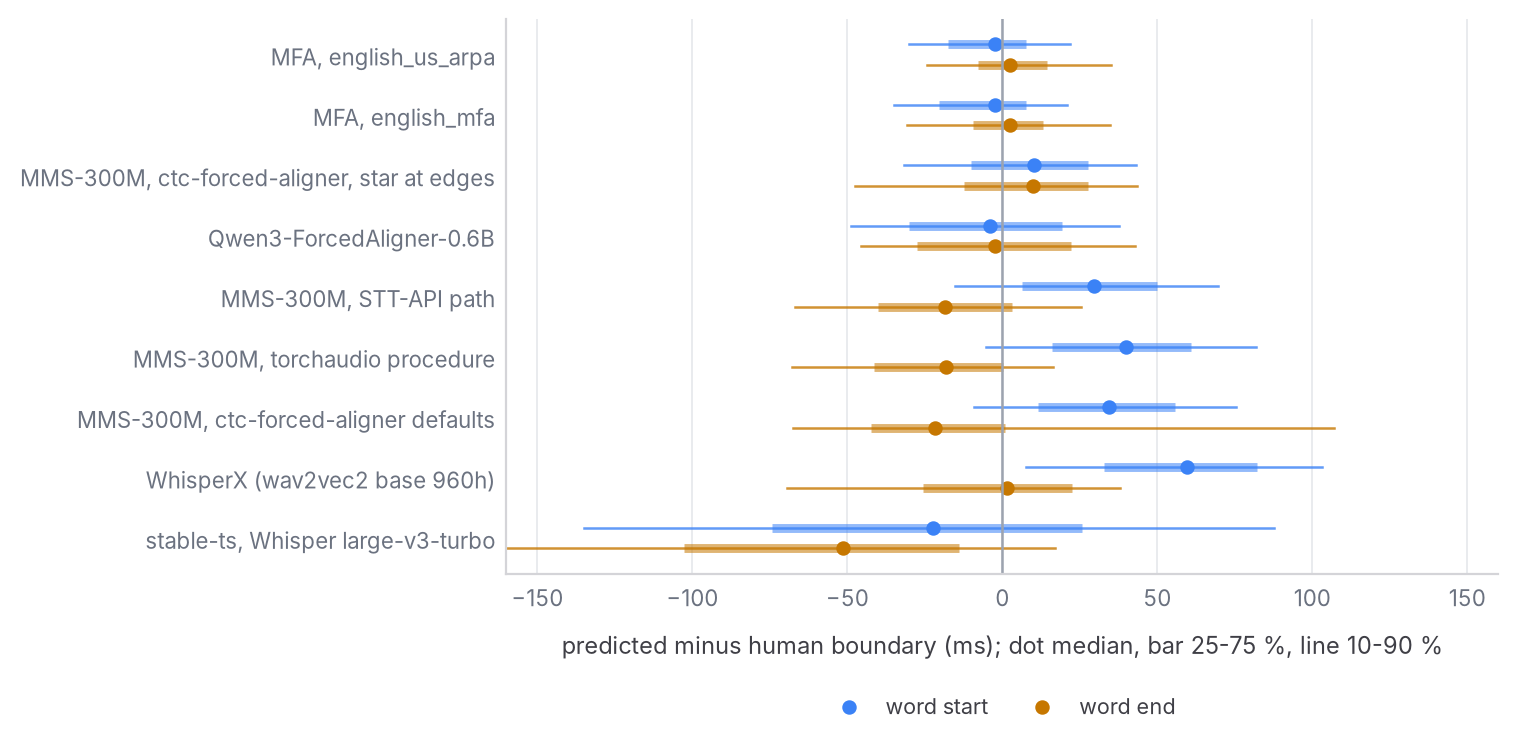}
\caption{Signed start and end errors on clean TIMIT: median (dot), interquartile range (thick bar) and 10 to 90\%
range (thin line). Positive is later than the human boundary.}
\label{fig:pilotbias}
\end{figure}

The CTC decoders are not just noisy, they are biased (Figure~\ref{fig:pilotbias}). With the torchaudio procedure,
words start a median \val{timit/mms-torchaudio/clean/sbias} ms late and end \val{timit/mms-torchaudio/clean/ebias} ms
early, the spike offsets of Section~\ref{sec:ctc}; WhisperX starts words \val{timit/whisperx/clean/sbias} ms late.
stable-ts errs the other way, ending words \val{timit/stablets/clean/ebias} ms early. MFA and Qwen3-ForcedAligner are
within a few milliseconds of zero at both ends. A bias is the one error a later stage can remove, and another reason
not to read boundaries off a CTC path unchanged.

\subsection{Telephone speech}

\begin{figure}[H]
\centering
\includegraphics[width=0.92\linewidth]{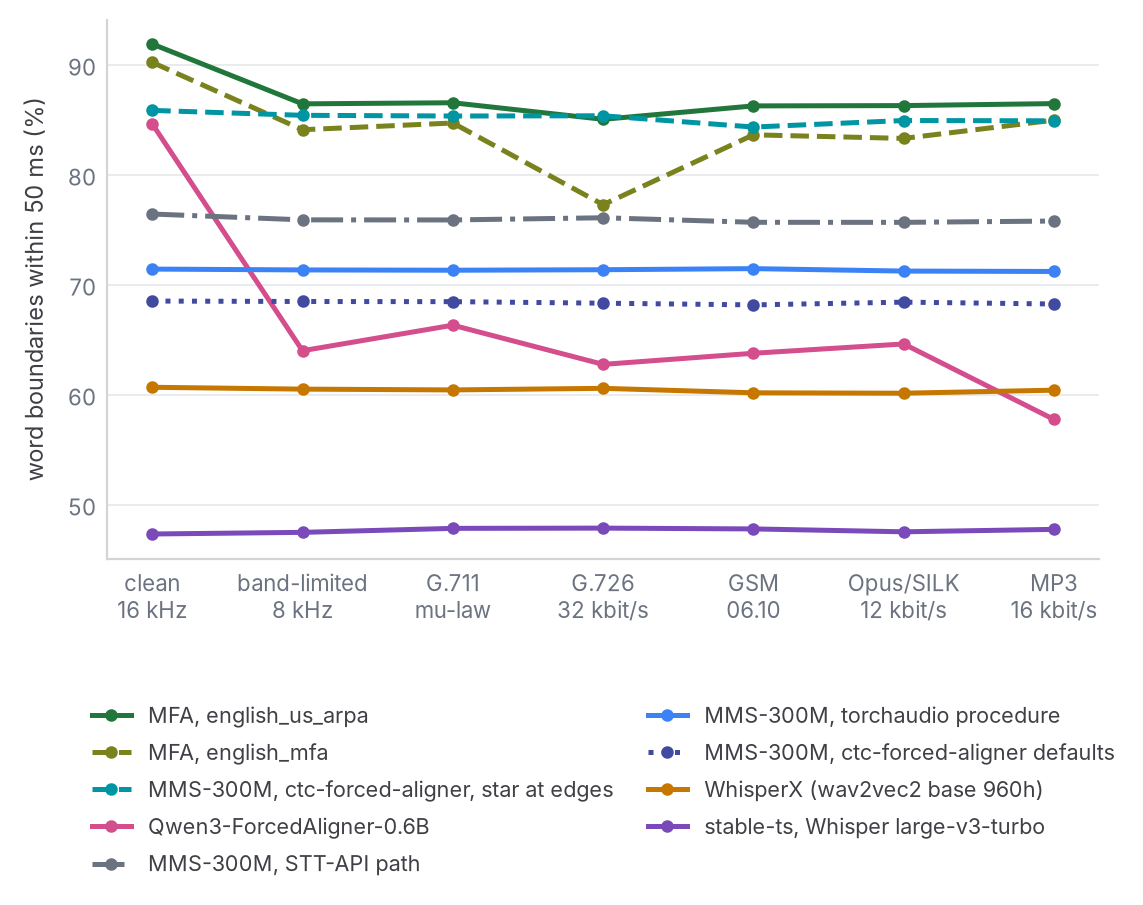}
\caption{Share of TIMIT word boundaries within 50 ms in each telephone condition of Table~\ref{tab:channel}.}
\label{fig:pilotchannel}
\end{figure}

Figure~\ref{fig:pilotchannel} follows each aligner through the telephone conditions, and two groups appear. The
precise aligners lose to bandwidth alone: band-limiting to 8 kHz costs MFA \val{timit/mfa-arpa/pcm/dw50} points of boundaries
within 50 ms and Qwen3-ForcedAligner \val{timit/qwen3fa/pcm/dw50}, whose mean error at 8 kHz, \val{timit/qwen3fa/pcm/mae} ms, is worse than all four
MMS decoders. The codecs on top cost little more, with two exceptions: MP3 takes Qwen3-ForcedAligner to
\val{timit/qwen3fa/mp3/mae} ms, and G.726 takes MFA's \texttt{english\_mfa} model to
\val{timit/mfa-english/g726/mae} ms. The CTC aligners barely move: MMS with the torchaudio procedure changes by \val{timit/mms-torchaudio/g711u/dmae} ms
of MAE under G.711. Table~\ref{tab:pilotchannel} gives the errors.

\begin{table}[H]
\centering
\caption{Mean absolute boundary error (ms) on TIMIT test in each telephone condition, scored on the utterances every aligner aligned in that condition. The best aligner per condition is in bold.}
\label{tab:pilotchannel}
\footnotesize
\begin{adjustbox}{max width=\linewidth}
\begin{tabular}{lrrrrrrr}
\toprule
\schead \scth{aligner} & \scth{clean} & \scth{8 kHz} & \scth{G.711} & \scth{G.726} & \scth{GSM} & \scth{Opus} & \scth{MP3} \\
\midrule
MFA, arpa & \textbf{19.1} & \textbf{25.3} & \textbf{25.3} & \textbf{26.8} & \textbf{25.8} & \textbf{25.5} & \textbf{25.4} \\
MFA, mfa & 21.3 & 28.1 & 27.9 & 37.2 & 29.5 & 29.2 & 27.3 \\
MMS, CFA edges & 29.2 & 29.8 & 29.8 & 29.8 & 30.9 & 30.2 & 30.4 \\
Qwen3-FA & 30.1 & 52.7 & 50.1 & 54.4 & 52.6 & 52.3 & 64.8 \\
MMS, STT-API & 35.6 & 36.5 & 36.4 & 36.6 & 37.2 & 37.1 & 36.9 \\
MMS, torchaudio & 37.5 & 37.9 & 37.9 & 37.9 & 38.2 & 38.1 & 38.1 \\
MMS, CFA default & 46.8 & 47.1 & 47.2 & 47.3 & 48.0 & 47.6 & 47.6 \\
WhisperX & 46.9 & 47.3 & 47.3 & 47.3 & 47.5 & 47.6 & 47.3 \\
stable-ts & 77.5 & 78.8 & 78.3 & 78.1 & 77.8 & 78.7 & 78.2 \\
\bottomrule
\end{tabular}
\end{adjustbox}
\end{table}

\subsection{Conversation and real telephone lines}

\begin{figure}[H]
\centering
\includegraphics[width=0.9\linewidth]{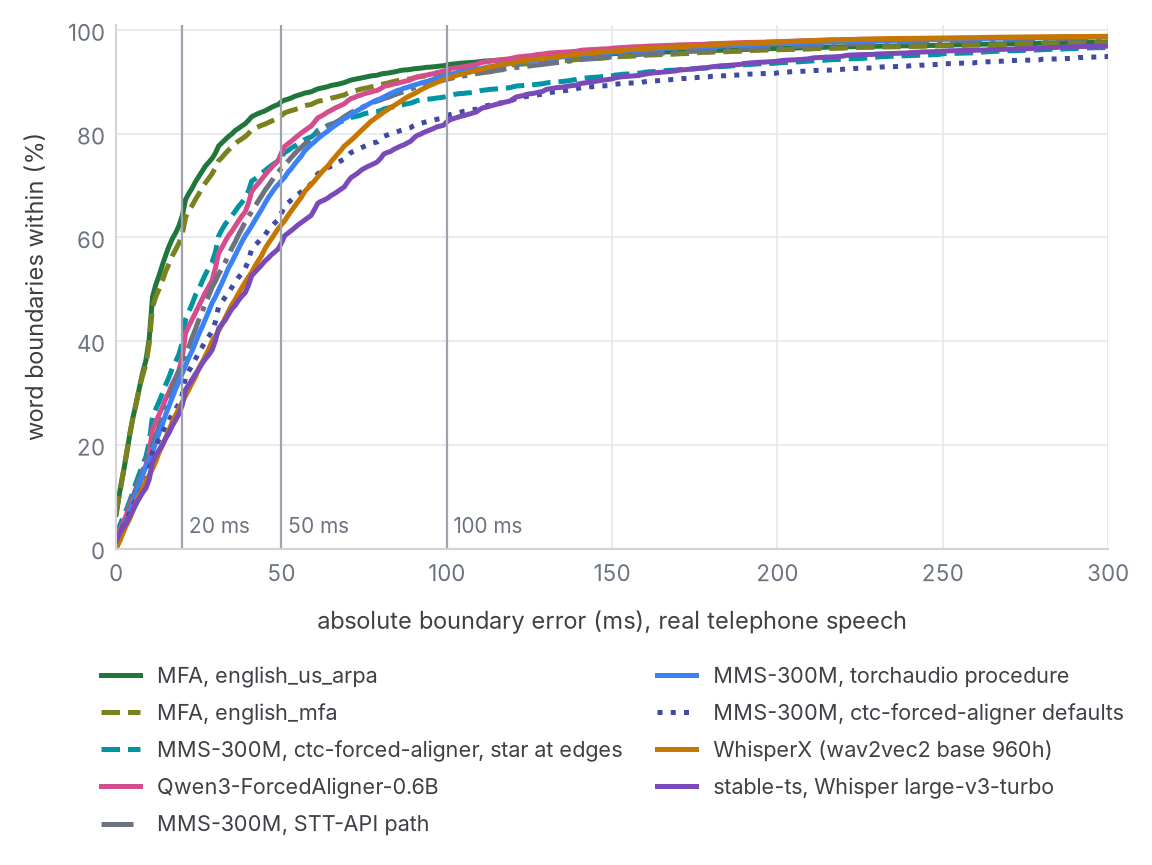}
\caption{Share of word boundaries within a given error on Switchboard conversations recorded over real landlines
(\val{swb/n-utt} utterances, \val{swb/n-words} words, hand-marked by the ICSI transcription project).}
\label{fig:pilotcdfswb}
\end{figure}

Table~\ref{tab:pilotsets} adds conversational speech and real telephone lines. On Buckeye the ranking mostly holds
(Qwen3-ForcedAligner moves ahead of the MMS decoders), and the package default stops being merely imprecise: on clean Buckeye its own span check rejected
\val{buckeye/cfa-segment/clean/fail-assert} utterances, and on another its native aligner crashed the process, though
not on every run of the clip. The torchaudio procedure, on the same emissions, never failed. On
Switchboard, conversations recorded over real landlines, MFA still places \val{swb/mfa-arpa/tel/w20}\% of boundaries
within 20 ms, but its mean error grows to \val{swb/mfa-arpa/tel/mae} ms: most boundaries stay precise and a few go
badly wrong. Figure~\ref{fig:pilotcdfswb} shows the tail.

\begin{table}[H]
\centering
\caption{Mean absolute word-boundary error (ms) on the three human-labelled sets: TIMIT read speech, Buckeye conversation, and Switchboard conversation recorded over real landlines (ICSI transcription project). G.711 columns are the simulated channel of Section~\ref{sec:channel}. The last column is the share of Switchboard boundaries within 20 ms. Best per column in bold.}
\label{tab:pilotsets}
\footnotesize
\begin{adjustbox}{max width=\linewidth}
\begin{tabular}{lrrrrrr}
\toprule
\schead \scth{aligner} & \scth{TIMIT} & \scth{+G.711} & \scth{Buckeye} & \scth{+G.711} & \scth{SWB} & \scth{SWB, 20 ms} \\
\midrule
MFA, arpa & \textbf{19.1} & \textbf{25.3} & \textbf{21.0} & \textbf{26.6} & \textbf{36.5} & \textbf{64.2} \\
MFA, mfa & 21.3 & 27.9 & 24.4 & 29.4 & 40.5 & 60.6 \\
MMS, CFA edges & 29.2 & 29.8 & 37.0 & 44.0 & 56.4 & 39.7 \\
Qwen3-FA & 30.1 & 50.1 & 30.9 & 42.2 & 42.3 & 36.9 \\
MMS, STT-API & 35.6 & 36.4 & 40.3 & 42.4 & 46.9 & 35.9 \\
MMS, torchaudio & 37.5 & 37.9 & 41.4 & 43.9 & 47.8 & 33.8 \\
MMS, CFA default & 46.8 & 47.2 & 43.3 & 47.1 & 71.6 & 29.7 \\
WhisperX & 46.9 & 47.3 & 50.2 & 50.6 & 53.7 & 28.2 \\
stable-ts & 77.5 & 78.3 & 61.3 & 61.9 & 67.2 & 27.6 \\
\bottomrule
\end{tabular}
\end{adjustbox}
\end{table}

\subsection{Speed}

Every aligner runs at least \val{timit/clean/speedup-min} times faster than real time on one shared GPU (real-time
factors \val{timit/clean/rtf-min} to \val{timit/clean/rtf-max} in Table~\ref{tab:pilot}). Speed will not decide the
design; precision will.

\section{Study design}
\label{sec:design}

\textbf{The bar.} MFA with its English ARPA model is the aligner to beat. On clean TIMIT its word boundaries miss by
\val{timit/mfa-arpa/clean/mae} ms on average and \val{timit/mfa-arpa/clean/median} ms at the median, with
\val{timit/mfa-arpa/clean/w20}\% within 20 ms. On the Buckeye speakers held out of training the mean is
\val{buckeye-held/mfa-arpa/clean/mae} ms, and on Switchboard's phone lines \val{swb/mfa-arpa/tel/mae} ms
(Table~\ref{tab:pilot}). A trained aligner counts as a result
only if it beats MFA on both the mean error and the share within 20 ms on every English set, clean and through every
codec, and beats MFA's own Korean and Dutch models on held-out speakers of the Seoul Corpus and the IFA corpus.

Three things put MFA ahead, and each becomes an axis. It segments at 10 ms, where Whisper and the CTC models answer on a
20 ms grid. It assigns every frame to a unit, silence included, so a word ends where its sound ends; a CTC aligner ends
it where its decoder chose to emit (Section~\ref{sec:ctc}). And its word edges are the edges of phones. Its weaknesses
are as plain: trained on clean wideband speech, it loses \val{timit/mfa-arpa/g711u/dmae} ms through G.711 where the best
CTC aligner loses \val{timit/cfa-edges/g711u/dmae}, and every language needs its own dictionary and acoustic model.

\textbf{Two families on one encoder.} The study compares two ways of reading boundaries off the same encoder. In the
CTC family the encoder emits character posteriors and Viterbi forced alignment reads the path; the head is also trained
on a frame target from the human word edges, inside a word or not, so that its edges stop following the spikes. The
language-model family starts from Whisper-Scicom, our Whisper large-v3 fine-tuned on our private dataset of
human-annotated Malaysian call-centre speech. Its decoder reads the transcript in Whisper's own timestamp format, at the
level of words. Whisper brackets a segment's text with a timestamp token on each side; here every word gets such a pair,
and the pair is two slots the decoder fills:
\begin{center}
\footnotesize
\texttt{<|startoftranscript|> <|en|> <|transcribe|>}\\
\texttt{<|startoflm|> his <|startofprev|> <|startoflm|> captain <|startofprev|> \dots{}}\\
\texttt{<|endoftext|>}
\end{center}
The slots are two special tokens Whisper already has, \texttt{<|startoflm|>} for a start and \texttt{<|startofprev|>}
for an end. They enter as token ids, so the byte-level tokenizer never splits them, and only the words are encoded as
text. The decoder runs once over the whole sequence, and its hidden state at each slot is scored against the
embeddings of Whisper's 1{,}501 timestamp tokens, \texttt{<|0.00|>} to \texttt{<|30.00|>} in 20~ms steps, to which
its output layer is tied. That gives every boundary a distribution over time in one pass. Training splits each human
boundary between its two neighbouring tokens (1.234~s puts 0.30 on \texttt{<|1.22|>} and 0.70 on \texttt{<|1.24|>}),
and decoding takes the expectation within two tokens of the most likely one, so the output is not tied to the grid.
No timestamp enters the input and none is fed back, so all words are timed at once and the text cannot change. The
times are made monotonic afterwards: a start never precedes the one before it, and an end never its own start.
Because the decoder is causal, a word's start slot is read before the word itself. Two variants test these choices
(Section~\ref{sec:slots}): the slots after the word instead of around it, and Whisper's format as Whisper is trained,
\begin{center}
\footnotesize
\texttt{<|startoftranscript|> <|en|> <|transcribe|>}\\
\texttt{<|0.56|> his <|0.76|> <|0.76|> captain <|1.26|> \dots{}}\\
\texttt{<|endoftext|>}
\end{center}
with real timestamp tokens rounded to 20~ms, next-token cross-entropy over the whole vocabulary, and the slots filled
in order at inference, each the most likely timestamp no earlier than the one before. Both families start from
Whisper-Scicom's encoder and differ in the head alone. A third arm fine-tunes MMS-300M, the model our own stack aligns
with today, with a CTC head. Each arm is trained once with LoRA and once in full, at one default each.

\textbf{Human labels only.} Every boundary the models train on was placed or corrected by people
(Table~\ref{tab:trainsets}): TIMIT's training speakers, Buckeye's speakers s05 to s20, the second phase of the
Switchboard transcription project, the Seoul Corpus of Korean, the IFA corpus of Dutch, and VoxAngeles' isolated words
in 95 languages. That is \val{fa/train-hours} hours and \val{fa/train-words} words in \val{fa/train-languages}
languages; DoReCo's 43 languages with audio join when its audio host answers again. We check a corpus's claim of
correction instead of taking it on trust. A forced aligner puts boundaries on a 10 ms grid and a person rarely does,
and the L2-ARCTIC subset documented as hand-aligned keeps \val{fa/grid/l2arctic}\% of its word boundaries there, so it
stays out. No machine alignment enters training, Whisper-Scicom's included. The pilot's test sets stay out, and so do
held-out speakers of every training corpus and \val{fa/unseen-languages} whole languages of VoxAngeles, which measure
transfer to a language the model never trained on. Stretches between pauses are short, so consecutive windows of one
speaker are also joined into examples of up to 30 s (\val{fa/joined-hours} hours): the calls the aligner serves come in
30 s windows.

\begin{table}[H]
\centering
\caption{The training data: every word boundary placed or corrected by people. Hours of audio in the windows the models see (at most 30 s); \emph{test}: held-out speakers or recordings; \emph{unseen}: whole languages held out; \emph{grid}: share of word boundaries exactly on a 10 ms grid, where a forced aligner puts them and a person rarely does; \emph{NC}: non-commercial. The pilot's English test sets are not in it.}
\label{tab:trainsets}
\scriptsize
\setlength{\tabcolsep}{4pt}
\begin{adjustbox}{max width=\linewidth}
\begin{tabular}{llllrrrrrrl}
\toprule
\schead \scth{corpus} & \scth{speech} & \scth{channel} & \scth{edges} & \scth{grid} & \scth{lang.} & \scth{train h} & \scth{words} & \scth{test h} & \scth{unseen h} & \scth{licence} \\
\midrule
TIMIT, train & read & clean & corrected & 4\% & 1 & 2.8 & 28{,}578 & -- & -- & LDC \\
Buckeye, s05--s20 & conversation & clean & corrected & 6\% & 1 & 8.7 & 119{,}438 & 1.4 & -- & NC \\
Switchboard, WS97 & conversation & phone & manual & 18\% & 1 & 2.1 & 26{,}231 & -- & -- & LDC \\
Seoul Corpus & interviews & clean & corrected & 0\% & 1 & 27.1 & 197{,}712 & 3.1 & -- & CC BY-NC \\
IFA & read, stories & clean & corrected & 0\% & 1 & 5.2 & 44{,}616 & 0.5 & -- & GPL-3.0 \\
VoxAngeles & single words & field & corrected & 24\% & 95 & 0.9 & 3{,}858 & 0.1 & 0.1 & CC BY-NC \\
L2-ARCTIC, suitcase & spont., L2 & clean & reviewed MFA & 89\% & 1 & \multicolumn{4}{l}{excluded: left on the aligner's grid} & CC BY-NC \\
\midrule
\textbf{total} &  &  &  &  & 97 & 46.7 & 420{,}433 & 5.0 & 0.1 &  \\
\bottomrule
\end{tabular}
\end{adjustbox}
\end{table}

\textbf{Adapters.} The winning family is swept on the grid of our hallucination study, one factor at a time around a
default: LoRA of rank 32 with $\alpha = 2r$ on the attention and MLP projections of encoder and decoder, at a
learning rate of $2\times10^{-4}$. The variants take rank 8 to 128, the attention projections alone, learning rates of
$10^{-4}$ and $5\times10^{-4}$, the hallucination study's $\alpha/r = 0.2$, and two or three times the training.
Alignment adds a factor that transcription did not need: adapters on the decoder alone, with the encoder frozen,
against adapters on both. Two more seeds of the default measure the spread a factor has to clear. The other family is
swept too only if it comes within two points of the winner. Output resolution, phone edges as a second target, and
telephone augmentation then follow on the winner (Table~\ref{tab:design}). For augmentation, AMR-NB and G.723.1 stay
out of training, so that the models are also tested on codecs they never heard.

\begin{table}[H]
\centering
\caption{The axes of the training ablation, with the stage that runs each. The head is the question the study is
built around; the adapter grid follows our hallucination study; resolution, phone edges and telephone augmentation
target what puts MFA ahead. Stages 2 to 5, the checkpoint, long recordings, transcript noise and the English-only
arm are reported in Section~\ref{sec:results}.}
\label{tab:design}
\footnotesize
\begin{tabular}{lL{5.2cm}L{3.9cm}c}
\toprule
\schead \scth{axis} & \scth{values} & \scth{question} & \scth{stage} \\
\midrule
head & CTC over characters with Viterbi, plus a frame target from human word edges; Whisper-Scicom's decoder predicting
word start and end timestamps; MMS-300M with CTC & which head places boundaries best from the same encoder? & 2 \\
fine-tuning & LoRA; full fine-tune & is LoRA enough? & 2 \\
adapter rank & LoRA $r$ of 8, 16, 32, 64, 128 with $\alpha = 2r$ & how much capacity does timing need? & 3 \\
adapter modules & attention projections; attention and MLP & does the MLP matter for timing? & 3 \\
adapter placement & decoder only; encoder and decoder & does the encoder have to move? & 3 \\
adapter scale & $\alpha/r$ of 2; 0.2 & does the ratio that protects transcription help? & 3 \\
learning rate & $10^{-4}$, $2\times10^{-4}$, $5\times10^{-4}$ & which rate converges without forgetting? & 3 \\
training length & 1{,}000, 2{,}000, 4{,}000, 6{,}000 steps of 16 clips; five seeds of the default & is the default
trained long enough, and how noisy is one run? & 3 \\
output resolution & the most likely 20 ms timestamp; the expectation within 1, 2 or 5 timestamps of it, or over
all; finer timestamps only if the grid binds & how much of the error is the grid? & 4 \\
supervision & word edges; plus an encoder head trained on word edges, or on word and phone edges (TIMIT, IFA, Seoul,
VoxAngeles) & do phone edges sharpen word edges? & 4 \\
slot order & start slot before the word (Whisper's order); both slots after it; Whisper's own layout with real
timestamp tokens & must the decoder see the word before timing it? & 4 \\
telephone augmentation & probability 0, 0.25, 0.5, 0.75, 1 over the six conditions of Table~\ref{tab:channel}, with
noise; AMR-NB and G.723.1 kept out of training & clean cost against telephone gain, and on unseen codecs & 5 \\
training languages & English sets; all human-labelled sets & does multilingual human data
transfer? & 2, 6 \\
checkpoint & Whisper-Scicom; Whisper large-v3; large-v3-turbo (4 decoder layers) & does the call-centre fine-tune
matter, and what does a small decoder cost? & 7 \\
long recordings & a coarse MMS-300M pass, then chunks of at most 25 s for the 30 s model & is precision kept on
minutes of audio? & 7 \\
transcript noise & 0 to 20\% inserted, deleted or substituted words & robustness to imperfect transcripts & 7 \\
\bottomrule
\end{tabular}
\end{table}

\textbf{Selection.} Among the configurations that clear the bar, the share of boundaries within 20 ms decides, pooled
over the human-labelled test sets with clean and telephone conditions weighted equally; the mean error breaks ties. No
candidate may lose more than two points between its clean and telephone conditions.

\section{Trained aligners}
\label{sec:results}

We trained both families on the human-labelled set and scored every model through the pilot's harness, on the pilot's
English sets and on held-out speakers and recordings of the Seoul, IFA and VoxAngeles corpora.

\subsection{Training}

Every run trains 2{,}000 steps of 16 clips, \val{s3/epochs-2k} of an epoch over the \val{fa/train-rows} training
examples (single windows and joined ones). Each step draws 60\% joined examples of up to 30 s. Half of the clips go
through one of the six telephone conditions of Table~\ref{tab:channel}, picked at random, with white noise at 10 to
40~dB SNR on half of those; the codec's delay is removed as in Section~\ref{sec:channel}, so the labels stay valid.
LoRA has rank 32 and $\alpha = 64$ on the attention and MLP projections of the encoder and the decoder, at a learning
rate of $2\times10^{-4}$, and trains \val{s3/ts-lora/params} million parameters. The full fine-tunes run at $10^{-5}$
($5\times10^{-5}$ for MMS-300M) on the same 16 clips per step. The CTC runs keep the encoder frozen for their first
200 steps while the new heads warm up. The timestamp model trains on the two-hot targets and decodes with the
expectation of Section~\ref{sec:design}. One clip in ten trains without a language token. Under
bf16, the MMS-300M runs diverged until updates with non-finite gradients were skipped.

\subsection{Two families}
\label{sec:families}

Table~\ref{tab:stage2} sets the six runs against the bar. Every one clears it: each beats MFA on both the mean error
and the share within 20~ms in all \val{s2/n-bar} set-conditions that have a bar, every English set clean and through
every codec, and Korean and Dutch against MFA's own models.

\begin{table}[H]
\centering
\caption{The two families against the bar (stage 2): mean absolute word-boundary error in ms / share of boundaries within 20 ms in \%, each set scored on the utterances every aligner aligned. Codec rows average the six conditions of Table~\ref{tab:channel} (TIMIT) or G.711, GSM and Opus (Buckeye). The bar is MFA with \texttt{english\_us\_arpa}, and MFA's own Korean and Dutch models on Seoul and IFA; VoxAngeles has none. \emph{CTC, Whisper-Scicom}: Whisper-Scicom's encoder with a CTC head; \emph{timestamps}: Whisper-Scicom's decoder predicting word timestamps. Every run trained 2{,}000 steps of 16 clips on all training languages. Bold: the best run per row on each measure. Every run beats the bar on both measures in all 16 set-conditions that have one.}
\label{tab:stage2}
\scriptsize
\setlength{\tabcolsep}{3pt}
\begin{adjustbox}{max width=\linewidth}
\begin{tabular}{lccccccc}
\toprule
\schead & & \multicolumn{2}{c}{\scth{CTC, MMS-300M}} & \multicolumn{2}{c}{\scth{CTC, Whisper-Scicom}} & \multicolumn{2}{c}{\scth{timestamps}} \\
\cmidrule(lr){3-4}\cmidrule(lr){5-6}\cmidrule(lr){7-8}
\schead \scth{set} & \scth{MFA bar} & \scth{LoRA} & \scth{full} & \scth{LoRA} & \scth{full} & \scth{LoRA} & \scth{full} \\
\midrule
TIMIT & 19.1/68.2 & 17.2/74.0 & 16.9/74.0 & 16.4/76.0 & 16.2/76.5 & \textbf{10.5}/\textbf{89.1} & 10.7/88.7 \\
TIMIT, six codecs & 25.7/62.7 & 17.2/73.6 & 16.9/74.2 & 16.4/75.8 & 16.4/75.7 & \textbf{11.1}/\textbf{88.0} & 11.4/87.4 \\
Buckeye s01--s04 & 21.5/70.7 & 15.7/77.9 & 15.4/77.6 & 14.0/80.6 & 13.9/80.5 & \textbf{12.1}/\textbf{85.8} & 12.4/85.5 \\
Buckeye, three codecs & 26.6/64.5 & 16.2/76.0 & 15.9/77.1 & 14.7/78.9 & 14.6/79.5 & \textbf{13.0}/\textbf{84.6} & 13.3/84.5 \\
Switchboard (telephone) & 36.5/64.2 & 22.6/74.7 & \textbf{21.8}/75.2 & 22.3/78.0 & 22.4/78.0 & 23.0/\textbf{80.7} & 22.7/80.5 \\
TIMIT, 20--28 s windows & 19.5/68.1 & 17.4/73.7 & 16.9/74.2 & 15.6/78.1 & 15.7/78.3 & \textbf{10.3}/\textbf{89.5} & 10.5/88.9 \\
Seoul (ko) & 21.3/70.2 & 16.2/81.0 & 15.6/81.7 & 13.5/86.0 & 14.0/85.7 & \textbf{11.3}/\textbf{90.0} & 11.4/89.9 \\
IFA (nl) & 335.4/38.1 & 18.6/73.1 & 17.8/73.5 & 18.1/74.9 & 17.4/75.5 & 15.3/\textbf{78.4} & \textbf{15.1}/78.2 \\
VoxAngeles & -- & 20.3/67.4 & 20.7/66.9 & 18.8/73.1 & 18.5/73.5 & 15.9/80.2 & \textbf{15.5}/\textbf{80.5} \\
\bottomrule
\end{tabular}
\end{adjustbox}
\end{table}

The timestamp model is the most precise. It places the most boundaries within 20~ms on every set,
\val{s2/ts-lead-min} to \val{s2/ts-lead-max} points ahead of the best CTC run, and it has the lowest mean error
everywhere except Switchboard. There the CTC runs lead (MMS-300M fine-tuned in full: \val{s2/swb/best-ctc-mae}~ms
against \val{s2/ts-lora/swb/mae}), while the timestamp model still places the most boundaries within 20~ms
(\val{s2/ts-lora/swb/w20}\% against at most \val{s2/swb/best-ctc-w20}\%). On clean TIMIT it nearly halves MFA's
error: \val{s2/ts-lora/timit/mae}~ms with \val{s2/ts-lora/timit/w20}\% of boundaries within 20~ms, against
\val{s2/bar/timit/mae}~ms and \val{s2/bar/timit/w20}\%. Through the six codecs its mean error rises to
\val{s2/ts-lora/timit-codecs/mae}~ms on average, and MFA's to \val{s2/bar/timit-codecs/mae}.

The head matters more than the encoder. With the same CTC head and recipe, Whisper-Scicom's encoder places
\val{s2/wenc-gain-min} to \val{s2/wenc-gain-max} points more boundaries within 20~ms than MMS-300M's, on every set.
With the same Whisper-Scicom encoder, predicting timestamps instead of reading a CTC path gains \val{s2/head-gain-min}
to \val{s2/head-gain-max} points. Full fine-tuning buys nothing: against LoRA it moves no set's mean error by more than
\val{s2/full-max-abs}~ms in either direction, so the sweep that follows is over LoRA alone.

These tests flatter every trained model in one respect: the English, Korean and Dutch test speakers come from
corpora the models trained on, so the models know those corpora's labelling conventions, and MFA does not.

\subsection{Against published aligners}
\label{sec:sota}

Whisper's own word timestamps, read off its cross-attention with dynamic time warping, sit far from human boundaries
(Table~\ref{tab:sota}): \val{timit/whisper-tm-cal/clean/mae}~ms on TIMIT for Whisper-Scicom even when its alignment
heads are the three of 640 that best fit human boundaries outside every test set, with their constant lag removed, and
96.8~ms through stable-ts under FA-Bench's protocol
\citep{chu2026fabench}. Trained on human word boundaries to read time off its timestamp tokens, the same model reaches
\val{s2/ts-lora/timit/mae}~ms in our harness, where MFA scores \val{timit/mfa-arpa/clean/mae}~ms.

\begin{table}[H]
\centering
\caption{Word boundaries on TIMIT against published aligners. \emph{Our harness}: all 1{,}344 SI and SX utterances of TIMIT test, every word's start and end; mean absolute error in ms / share within 20 ms in \%. \emph{FA-Bench}: TIMIT's 192-utterance core test, each word junction once, its utterance edges included; mean absolute error in ms / F1 at 20 ms \citep{chu2026fabench}. Published rows are quoted from FA-Bench's Table~1; whisper's own timestamps there are stable-ts, which aligns the reference text through Whisper's cross-attention (Whisper large-v3 decoding its own words scores 148.0~ms / 0.10). Ours under FA-Bench are scored with FA-Bench's own code, as means over the seeds listed, the same runs in both columns.}
\label{tab:sota}
\scriptsize
\setlength{\tabcolsep}{4pt}
\begin{adjustbox}{max width=\linewidth}
\begin{tabular}{llrrr}
\toprule
\schead \scth{aligner} & \scth{word times from} & \scth{seeds} & \scth{our harness} & \scth{FA-Bench} \\
\midrule
Whisper's own word timestamps & cross-attention + DTW &  & 41.2 / 41.5 & 96.8 / 0.18 \\
MFA 3.4 & HMM-GMM, English ARPA &  & 19.1 / 68.2 & 21.8 / 0.64 \\
MAPS & frame classifier &  & -- & 25.5 / 0.71 \\
Olign (commercial API) & not published &  & -- & 18.2 / 0.78 \\
\midrule
ours, CTC on MMS-300M & LoRA, CTC forced path &  & 17.2 / 74.0 & 19.2 / 0.76 \\
ours, CTC on Whisper-Scicom & the same &  & 16.4 / 76.0 & 18.2 / 0.79 \\
ours, timestamp tokens & LoRA, timestamp tokens & 5 & 10.5 / 89.1 & 11.6 / 0.89 \\
\quad slots after the word & the same & 5 & 10.5 / 89.0 & 11.6 / 0.90 \\
\quad the same on large-v3-turbo & the same & 2 & 10.4 / 89.0 & 11.5 / 0.90 \\
\bottomrule
\end{tabular}
\end{adjustbox}
\end{table}

The two protocols differ in their test utterances and in how they count boundaries, so we also ran our models
through FA-Bench's own code on its TIMIT splits (Table~\ref{tab:sota}, right). There the timestamp model scores
\val{fb/ts/mae}~ms at an F1 of \val{fb/ts/f1} within 20~ms (the mean of \val{fb/ts/n} seeds, which range from
\val{fb/ts/mae-lo} to \val{fb/ts/mae-hi}~ms). That is \val{fb/ts/cut-pct}\% under the best published error, the
commercial Olign API's 18.2~ms at an F1 of 0.78, and MFA 3.4 scores 21.8~ms \citep{chu2026fabench}. Our CTC model on
Whisper-Scicom's encoder already matches Olign, at \val{fb/ctcw/mae}~ms and \val{fb/ctcw/f1}. On FA-Bench's
development split the timestamp model scores \val{fb/ts/dev/mae}~ms. FA-Bench's Buckeye splits are cut at the
original corpus's silence and noise markers, which our segmented mirror lacks. Like MAPS, our models trained on TIMIT's training speakers,
so they know its labelling conventions; MFA's English model did not, and FA-Bench states only that Olign did not
train on its test splits.

\subsection{Transfer from English}
\label{sec:transfer}

How much of this needs the other languages? We trained the same timestamp recipe on the English rows alone (TIMIT,
Buckeye and Switchboard: \val{fa/en-rows} of the \val{fa/train-rows} examples) and scored it everywhere
(Table~\ref{tab:transfer}). On Korean, Dutch and VoxAngeles it meets languages, and labelling conventions, it never
saw. Whisper-Scicom's own word timestamps, the cross-attention aligner of Table~\ref{tab:sota} before any alignment
training, mark where that training starts.

\begin{table}[H]
\centering
\caption{How far timestamp training on English alone transfers: the stage-2 timestamp LoRA recipe trained on the English rows only, against the same recipe on all training languages, and untrained Whisper-Scicom (its word timestamps read off the cross-attention of three alignment heads chosen on human boundaries outside every test set). Mean absolute error in ms / share within 20 ms in \%, on each set's common utterances. The bar is MFA (\texttt{english\_us\_arpa}; its own Korean and Dutch models on Seoul and IFA). Seoul has 4 test speakers, IFA 1, and VoxAngeles one per language. $\dagger$: the untrained aligner needs a Whisper language token and aligned only the 439 of 1{,}055 VoxAngeles clips whose language has one. Bold: the best per row on each measure.}
\label{tab:transfer}
\scriptsize
\setlength{\tabcolsep}{4pt}
\begin{adjustbox}{max width=\linewidth}
\begin{tabular}{lrrrr}
\toprule
\schead \scth{set} & \scth{MFA bar} & \scth{untrained} & \scth{English only} & \scth{multilingual} \\
\midrule
TIMIT & 19.1/68.2 & 41.2/41.5 & 10.8/89.0 & \textbf{10.5}/\textbf{89.1} \\
TIMIT, G.711 & 25.3/63.1 & 41.6/40.9 & 11.1/88.3 & \textbf{10.8}/\textbf{88.4} \\
Buckeye s01--s04 & 21.5/70.7 & 36.8/40.1 & 12.1/85.2 & \textbf{12.1}/\textbf{85.8} \\
Switchboard (telephone) & 36.5/64.2 & 54.6/39.4 & 27.1/78.5 & \textbf{23.0}/\textbf{80.7} \\
TIMIT, 20--28 s windows & 19.5/68.1 & 50.4/39.2 & 10.4/89.2 & \textbf{10.3}/\textbf{89.5} \\
\midrule
Seoul (ko) & 21.3/70.2 & 87.2/37.8 & 24.2/79.1 & \textbf{11.3}/\textbf{90.0} \\
IFA (nl) & 335.4/38.1 & 167.3/27.7 & 22.2/71.7 & \textbf{15.3}/\textbf{78.4} \\
VoxAngeles, 90 languages & -- & 72.3/18.7$^\dagger$ & 28.3/61.7 & \textbf{15.9}/\textbf{80.2} \\
\quad 80 in the multilingual training & -- & 75.6/19.1$^\dagger$ & 25.7/63.1 & \textbf{14.3}/\textbf{82.0} \\
\quad 10 held out of all training & -- & 69.9/18.4$^\dagger$ & 30.3/60.6 & \textbf{17.2}/\textbf{78.9} \\
\bottomrule
\end{tabular}
\end{adjustbox}
\end{table}

\textbf{In English the other languages cost nothing.} The English-only and the multilingual model are within
\val{tr/english-max-diff}~ms of each other on TIMIT and Buckeye. On Switchboard the multilingual model is
\val{tr/swb/diff-mae}~ms and \val{tr/swb/diff-w20} points better, and fewer of its boundaries are off by more than
half a second (\val{swb/ft-ts-lora/tel/over500}\% against \val{swb/ft-ts-lora-en/tel/over500}\%).

\textbf{Most of the skill transfers.} On the utterances all three aligned, English-only training closes
\val{tr/closes/seoul/mae}\% (Korean), \val{tr/closes/ifa/mae}\% (Dutch) and \val{tr/closes/vox/mae}\% (VoxAngeles,
languages with a Whisper token) of the gap in mean error between the untrained aligner and the multilingual model, and
\val{tr/closes/seoul/w20}\%, \val{tr/closes/ifa/w20}\% and \val{tr/closes/vox/w20}\% of the gap within 20~ms. On
Korean it beats MFA's own Korean model at the median (\val{seoul/ft-ts-lora-en/clean/median} against
\val{seoul/mfa-korean/clean/median}~ms) and within 20~ms (\val{seoul/ft-ts-lora-en/clean/w20}\% against
\val{seoul/mfa-korean/clean/w20}\%), but not on the mean (\val{seoul/ft-ts-lora-en/clean/mae} against
\val{seoul/mfa-korean/clean/mae}~ms): \val{seoul/ft-ts-lora-en/clean/over500}\% of its boundaries are off by more than
half a second, against MFA's \val{seoul/mfa-korean/clean/over500}\%. It misses the bar there and nowhere else: it
beats MFA in \val{tr/en-beats} of the \val{tr/en-n-bar} set-conditions.

\begin{figure}[H]
\centering
\includegraphics[width=0.88\linewidth]{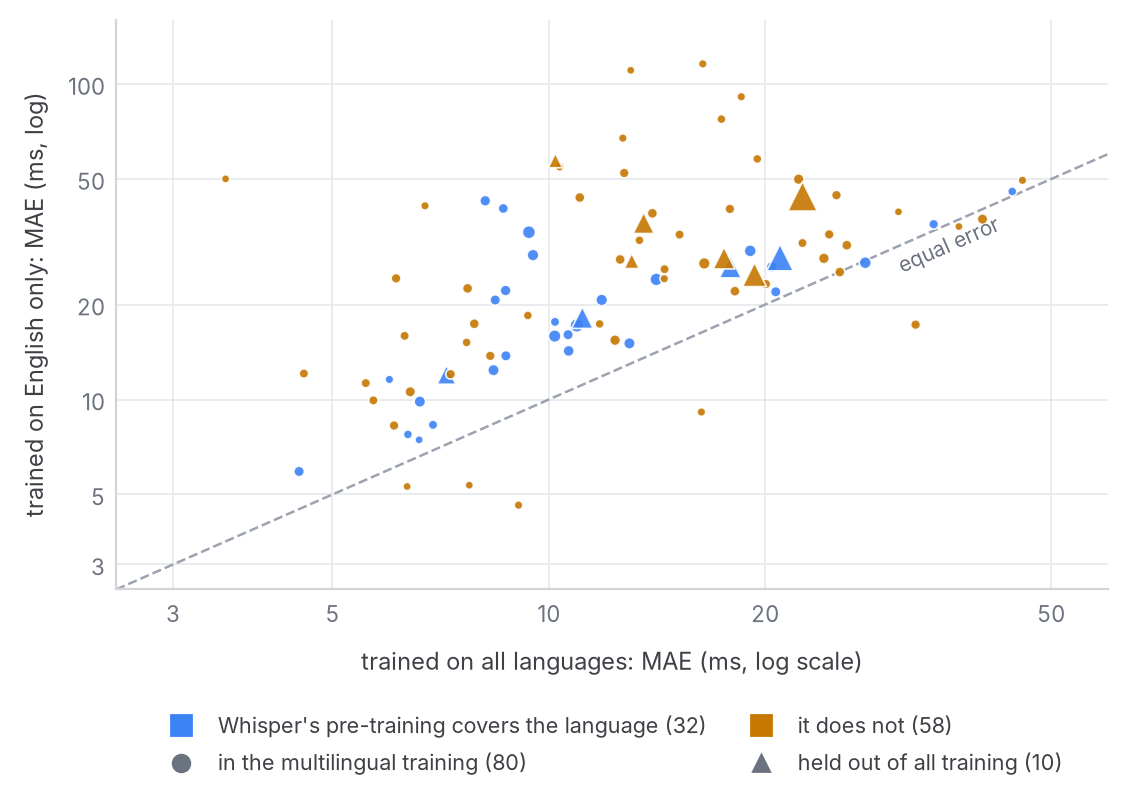}
\caption{Each VoxAngeles language's mean boundary error with timestamp training on English alone against training on
all languages; points above the diagonal are languages where the multilingual model is better. Colour: whether
Whisper's own pre-training covers the language (a language token, or a variety of one). Triangles: the languages held
out of all training. Marker area grows with the language's number of boundaries (one speaker each).}
\label{fig:transfer}
\end{figure}

\textbf{Training on many languages still cuts the error on them by \val{tr/cut-min} to \val{tr/cut-max}\%.} Korean
goes from \val{tr/seoul/en/mae} to \val{tr/seoul/multi/mae}~ms, Dutch from \val{tr/ifa/en/mae} to
\val{tr/ifa/multi/mae}, VoxAngeles from \val{tr/vox/en/mae} to \val{tr/vox/multi/mae}. The multilingual model is the
better of the two on \val{tr/vox/better-mae} of the \val{tr/vox/n-lang} VoxAngeles languages by mean error
(Figure~\ref{fig:transfer}).

\textbf{The transfer is better where Whisper's pre-training heard the language.} On the \val{tr/cov/covered/n}
VoxAngeles languages that Whisper's pre-training covers, the English-only model trails the multilingual one by
\val{tr/cov/covered/gap}~ms (\val{tr/cov/covered/en} against \val{tr/cov/covered/multi}); on the
\val{tr/cov/not-covered/n} it does not cover, by \val{tr/cov/not-covered/gap}~ms (\val{tr/cov/not-covered/en} against
\val{tr/cov/not-covered/multi}). The lag doubles: the difference, \val{tr/cov/diff}~ms, has a 95\% bootstrap
interval over languages of \val{tr/cov/diff-lo} to \val{tr/cov/diff-hi}~ms. Within 20~ms the English-only model falls
\val{tr/cov/w20-behind} points further behind, an interval of \val{tr/cov/w20-behind-lo} to
\val{tr/cov/w20-behind-hi} that includes zero. Covered and uncovered languages may differ in other ways too; the
multilingual model, as each group's control, absorbs only what affects both models alike.

\textbf{Training on the language itself adds nothing measurable once many languages are in.} The multilingual model
leads the English-only one by \val{tr/seen/seen/gap}~ms on the \val{tr/seen/seen/n} VoxAngeles languages it trained on
and by \val{tr/seen/unseen/gap}~ms on the \val{tr/seen/unseen/n} it never saw. If training on a language helped, the
first lead would be the larger; the difference is \val{tr/seen/diff}~ms (\val{tr/seen/diff-lo} to
\val{tr/seen/diff-hi}). With \val{tr/seen/unseen/n} held-out languages the interval is wide: it bounds the effect
rather than ruling it out.

\textbf{The language token does not matter.} Dropping it moves either model by at most \val{tr/nolang-max}~ms, with
one exception. The English-only model, which never trained with \texttt{<|nl|>}, does worse on Dutch without it: its
mean error goes from \val{ifa/ft-ts-lora-en/clean/mae} to \val{ifa/ft-ts-lora-en-nolang/clean/mae}~ms, and its share of
boundaries off by more than half a second from \val{ifa/ft-ts-lora-en/clean/over500}\% to
\val{ifa/ft-ts-lora-en-nolang/clean/over500}\%.

\begin{figure}[H]
\centering
\includegraphics[width=\linewidth]{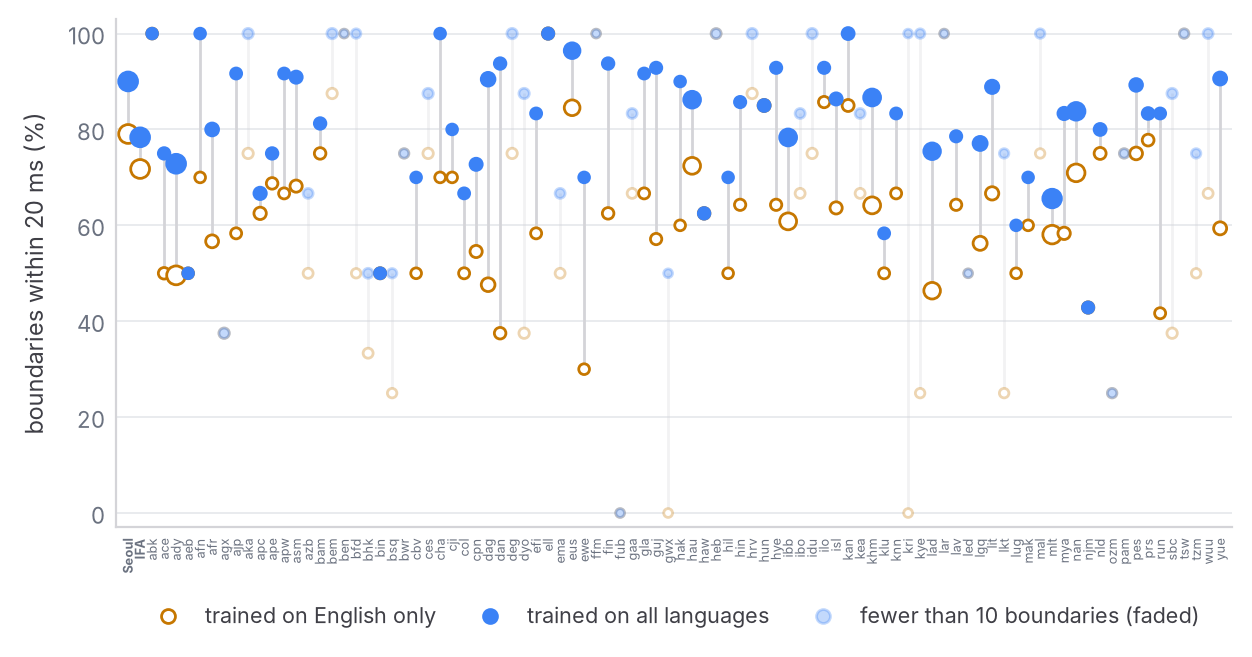}
\caption{Every language of the transfer test: the share of boundaries within 20~ms for the recipe trained on English
alone (hollow) and on all training languages (filled). \emph{Seoul} and \emph{IFA}, the Korean and Dutch held-out
speakers, come first; the VoxAngeles languages follow in alphabetical order of their ISO 639-3 codes, listed in
Table~\ref{tab:transferlang}. Marker area grows with the number of boundaries behind a language, and the
\val{trl/n-under10} languages with fewer than 10 are faded: their share moves in steps of ten points or more.}
\label{fig:transferlang}
\end{figure}

\textbf{Language by language.} Figure~\ref{fig:transferlang} puts every language side by side, before and after
the other languages join training, and Table~\ref{tab:transferlang} in the appendix lists them all. Training on all
languages raises the share of boundaries within 20~ms on \val{trl/better-w20} of the \val{trl/n-vox} VoxAngeles
languages, by a median of \val{trl/median-dw20} points, leaves \val{trl/same-w20} unchanged and lowers none. Korean
goes from \val{trl/kor/en-w20}\% to \val{trl/kor/all-w20}\%, and Dutch from \val{trl/nld/en-w20}\% to
\val{trl/nld/all-w20}\%. Most VoxAngeles rows rest on a handful of boundaries: half the languages have
\val{trl/median-bounds} or fewer, and a language with a single clip has only two, so its share can only be 0, 50 or
100\%, which is why some rows jump from one end to the other. The \val{trl/n-big} languages with at least 40 boundaries
gain a median of \val{trl/big-median-dw20} points, and none loses.

\textbf{Caveats.} VoxAngeles transcripts are IPA and each clip is one word, neither of which the English-only model
ever saw, so its VoxAngeles gap mixes language with transcript convention. Korean and Dutch, in their own spelling and
in running speech, are the cleaner test of language transfer. On Seoul and IFA the multilingual model trained on other
speakers of the same corpus, so part of its lead is their labelling convention. Korean is
\val{tr/matched/seoul/speakers} speakers and Dutch \val{tr/matched/ifa/speakers}.

\subsection{Adapter sweep}

Stage 3 sweeps the timestamp model alone: it led the CTC family by at least \val{s2/ts-lead-min} points within 20~ms
on every set, more than the two points under which both are swept. Every run is the whole encoder-decoder model with
the default recipe and one factor changed (Table~\ref{tab:sweep}). A factor counts when it moves the mean error over
the seven test sets by more than two standard deviations of \val{s3/n-seeds} seeds of the default.

\begin{table}[H]
\centering
\caption{The adapter sweep on the timestamp model (stage 3). Each run is the default (LoRA rank 32, $\alpha = 2r$, on attention and MLP of encoder and decoder, lr $2\times10^{-4}$, 2{,}000 steps of 16 clips) with one factor changed. Set columns: mean absolute error in ms on each set's common utterances (codecs: the mean of TIMIT's six). \emph{Mean}: MAE and share within 20 ms averaged over the seven sets in their own condition (codecs excluded). $z$: the mean MAE's distance from the seeds' mean in seed standard deviations (5 seeds: 14.00 ms, SD 0.12); positive is worse, bold beyond 2. \emph{Trainable}: LoRA parameters, millions. Bold set cells: the lowest error in the column.}
\label{tab:sweep}
\scriptsize
\setlength{\tabcolsep}{3pt}
\begin{adjustbox}{max width=\linewidth}
\begin{tabular}{lrrrrrrrrrrrr}
\toprule
\schead & & \multicolumn{8}{c}{\scth{MAE, ms}} & \multicolumn{2}{c}{\scth{mean}} & \\
\cmidrule(lr){3-10}\cmidrule(lr){11-12}
\schead \scth{run} & \scth{trainable} & \scth{TIMIT} & \scth{codecs} & \scth{Buckeye} & \scth{SWB} & \scth{20–28 s} & \scth{Seoul} & \scth{IFA} & \scth{VoxAng.} & \scth{MAE} & \scth{20 ms} & \scth{$z$} \\
\midrule
default (seed 0) & 58 & 10.5 & 11.1 & 12.1 & 23.0 & 10.3 & 11.3 & 15.3 & 15.9 & 14.06 & 84.82 &  \\
seed 1 & 58 & 10.5 & 11.2 & 12.1 & 23.6 & 10.5 & 11.5 & 15.3 & 15.5 & 14.15 & 84.91 &  \\
seed 2 & 58 & 10.5 & 11.2 & 12.2 & 23.2 & 10.2 & 11.5 & 14.9 & 15.3 & 13.97 & 85.07 &  \\
seed 3 & 58 & 10.4 & 11.1 & \textbf{11.9} & 23.1 & 10.2 & 11.4 & 14.9 & \textbf{14.8} & 13.82 & 84.97 &  \\
seed 4 & 58 & 10.6 & 11.3 & 12.0 & 23.1 & 10.2 & 11.5 & 15.5 & 15.1 & 14.00 & 84.80 &  \\
\midrule
rank 8 & 14 & 11.0 & 11.8 & 12.3 & 22.6 & 10.8 & 11.4 & 14.7 & 16.6 & 14.19 & 84.44 & $+1.6$ \\
rank 16 & 29 & 10.8 & 11.4 & 12.1 & 23.1 & 10.5 & 11.3 & 14.8 & 16.0 & 14.08 & 84.73 & $+0.7$ \\
rank 64 & 115 & 10.3 & 10.9 & 12.3 & 24.8 & 10.1 & \textbf{11.1} & 15.0 & 15.3 & 14.13 & 84.75 & $+1.1$ \\
rank 128 & 231 & 10.2 & 10.8 & 12.2 & 23.9 & 10.1 & 11.4 & 15.0 & 15.1 & 13.97 & 84.87 & $-0.3$ \\
\midrule
attention only & 31 & 11.0 & 11.5 & 12.0 & 23.1 & 10.7 & 11.3 & 14.9 & 16.2 & 14.17 & 84.52 & $+1.4$ \\
decoder only & 34 & 12.3 & 12.9 & 13.3 & 24.2 & 12.0 & 12.7 & 16.6 & 17.6 & 15.53 & 81.49 & $\mathbf{+12.8}$ \\
\midrule
lr $10^{-4}$ & 58 & 10.9 & 11.7 & 12.2 & 22.9 & 10.8 & 11.4 & 14.8 & 16.0 & 14.14 & 84.51 & $+1.2$ \\
lr $5\times10^{-4}$ & 58 & \textbf{10.1} & \textbf{10.7} & 12.1 & 24.0 & \textbf{10.0} & 11.6 & 14.9 & 15.7 & 14.04 & 85.00 & $+0.3$ \\
\midrule
$\alpha/r = 0.2$ & 58 & 11.5 & 12.3 & 12.4 & \textbf{22.2} & 11.2 & 11.6 & 15.3 & 16.9 & 14.44 & 84.00 & $\mathbf{+3.7}$ \\
\midrule
1{,}000 steps & 58 & 11.0 & 12.0 & 12.2 & 22.7 & 10.9 & 11.4 & \textbf{14.7} & 15.9 & 14.11 & 84.49 & $+0.9$ \\
4{,}000 steps & 58 & 10.4 & 10.8 & 12.2 & 24.5 & 10.3 & 11.8 & 15.1 & 15.6 & 14.28 & 84.36 & $\mathbf{+2.4}$ \\
6{,}000 steps & 58 & 10.3 & 10.9 & 12.3 & 24.7 & 10.4 & 12.0 & 15.7 & 15.6 & 14.43 & 83.87 & $\mathbf{+3.6}$ \\
\bottomrule
\end{tabular}
\end{adjustbox}
\end{table}

\textbf{Nothing beats the default.} Its \val{s3/n-seeds} seeds average \val{s3/seed-mean-mae}~ms (SD \val{s3/seed-sd-mae}) and
\val{s3/seed-mean-w20}\% within 20~ms (SD \val{s3/seed-sd-w20}). The best changed run, \val{s3/best/name}, scores
\val{s3/best/mae}~ms, \val{s3/best/z} SD better than the seeds' mean and inside their range (\val{s3/seed-mae-lo} to
\val{s3/seed-mae-hi}~ms). By the share within 20~ms, which the selection rule puts first, no run rises more than
two SDs above the seeds either (the best is \val{s3/max-z-w20} SD), and besides the three settings below, rank 8,
attention-only adapters and the lower learning rate fall more than two SDs below them. There are no winners to
combine, and the default stays.

\textbf{The encoder has to be adapted.} With adapters on the decoder alone and the encoder frozen, every set is worse,
by \val{s3/dec/dmae-min} to \val{s3/dec/dmae-max}~ms of mean error and \val{s3/dec/dw20-min} to
\val{s3/dec/dw20-max} points within 20~ms. Where words begin and end has to be learnt in the encoder, not only read
out by the decoder.

\textbf{Weaker updates cost a little.} At $\alpha/r = 0.2$, the ratio that kept our hallucination study's models from
forgetting the end of text, the adapters' output is a tenth of the default's. It acts like a much lower learning rate:
\val{s3/ts-a02/mean-mae}~ms and \val{s3/ts-a02/mean-w20}\%, \val{s3/ts-a02/z-mae} SD worse than the seeds. Here the
decoder only fills in timestamps between words it is given, so there is no end of text to forget.

\textbf{Training longer hurts, and steadily.} 4{,}000 steps (\val{s3/epochs-4k} epochs) is \val{s3/ts-4k/d-mae}~ms
and \val{s3/ts-4k/d-w20} points worse than the seeds' mean (\val{s3/ts-4k/z-mae} SD); 6{,}000 steps
(\val{s3/epochs-6k} epochs) is \val{s3/ts-6k/d-mae}~ms and \val{s3/ts-6k/d-w20} points worse (\val{s3/ts-6k/z-mae}
SD). TIMIT keeps improving a little (\val{s3/ts-4k/timit/mae} and \val{s3/ts-6k/timit/mae}~ms) while Switchboard
(\val{s3/ts-4k/swb/mae} and \val{s3/ts-6k/swb/mae}~ms), Korean and Dutch get worse. Half the training is no better: 1{,}000 steps (\val{s3/epochs-1k} of an
epoch) scores \val{s3/ts-1k/mean-mae}~ms and \val{s3/ts-1k/mean-w20}\% within 20~ms, worse on TIMIT
(\val{s3/ts-1k/timit/mae}~ms) and better on Switchboard (\val{s3/ts-1k/swb/mae}~ms), the trade that weaker adaptation
makes below.

\begin{figure}[H]
\centering
\includegraphics[width=0.88\linewidth]{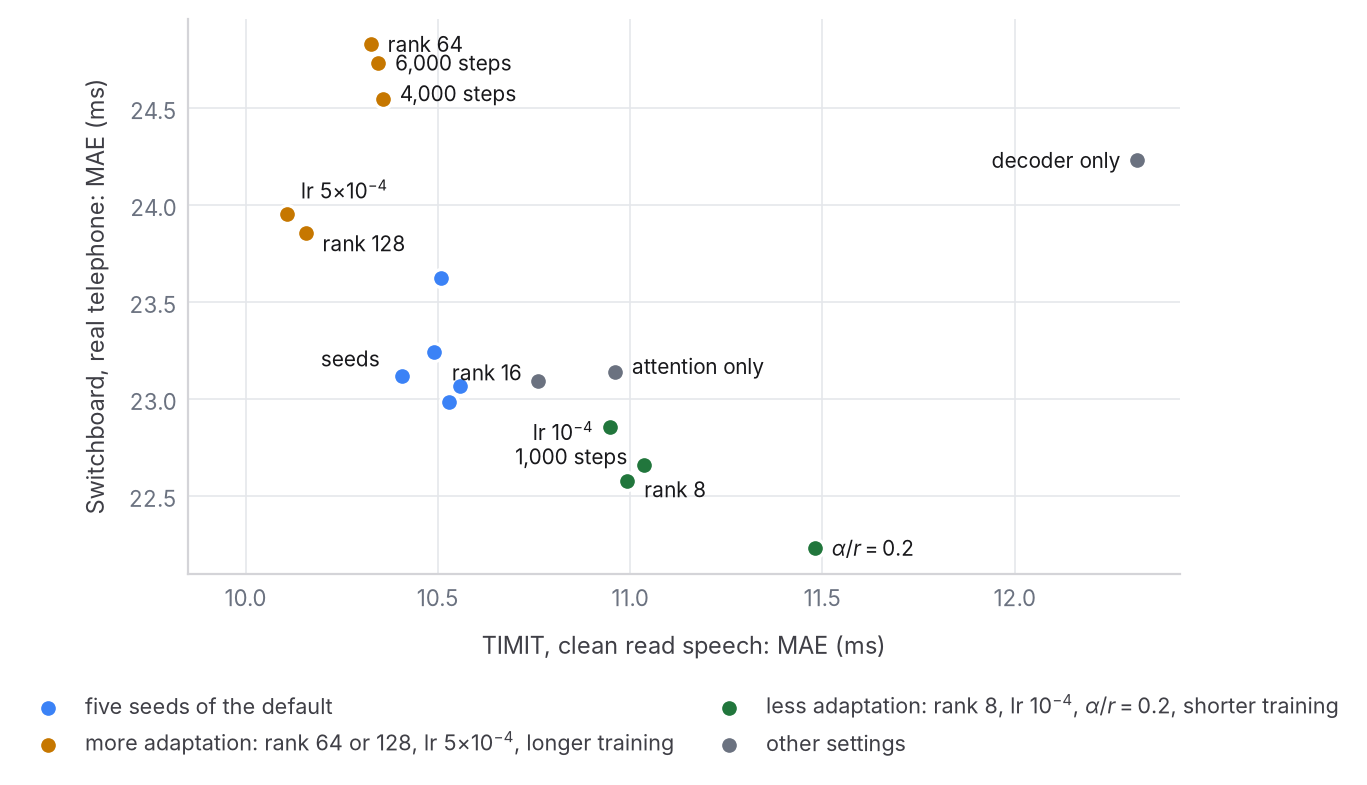}
\caption{Each run of the adapter sweep by its mean boundary error on clean TIMIT and on Switchboard's real telephone
calls. Lower left is better on both. More adaptation moves a run up and to the left, less moves it down and to the
right; adapting the decoder alone is worse on both.}
\label{fig:sweep}
\end{figure}

\textbf{How strongly to adapt trades TIMIT against Switchboard} (Figure~\ref{fig:sweep}). More adaptation (rank 64 or
128, a learning rate of $5\times10^{-4}$, 4{,}000 steps) lowers TIMIT's error to \val{s3/stronger/timit/lo} to
\val{s3/stronger/timit/hi}~ms, against \val{s3/seed-range/timit/lo} to \val{s3/seed-range/timit/hi} for the seeds, and
raises Switchboard's to
\val{s3/stronger/swb/lo} to \val{s3/stronger/swb/hi}~ms, against \val{s3/seed-range/swb/lo} to
\val{s3/seed-range/swb/hi} for the seeds. Less adaptation (rank 8, $10^{-4}$, $\alpha/r = 0.2$, 1{,}000 steps) does
the reverse:
TIMIT \val{s3/weaker/timit/lo} to \val{s3/weaker/timit/hi}~ms, Switchboard \val{s3/weaker/swb/lo} to
\val{s3/weaker/swb/hi}. The two sets move in opposite directions, so the mean barely moves. Switchboard, the one test
on real telephone lines, is \val{fa/swb-train-hours} of the \val{fa/train-hours-exact} training hours.

\textbf{Capacity hardly matters.} From rank 8 to rank 128, \val{s3/rank/params-lo} to \val{s3/rank/params-hi} million
trainable parameters, the mean error stays between \val{s3/rank/mean-lo} and \val{s3/rank/mean-hi}~ms.
Attention-only adapters (\val{s3/ts-attn/params} million) score close to rank 16 on attention and MLP
(\val{s3/ts-r16/params} million): \val{s3/ts-attn/mean-mae} against \val{s3/ts-r16/mean-mae}~ms.

\textbf{Seed noise differs by set.} TIMIT varies by \val{s3/seed-range/timit/width}~ms between seeds, so its moves
of a few tenths of a millisecond are real; Switchboard varies by \val{s3/seed-range/swb/width}~ms, VoxAngeles by
\val{s3/seed-range/vox/width} and IFA by \val{s3/seed-range/ifa/width}, so one run's differences of that size there
are noise. Every run, decoder-only included, still beats MFA in all \val{s3/n-bar} set-conditions.

\subsection{Output resolution}
\label{sec:resolution}

The timestamp model reads time off Whisper's 20~ms timestamp tokens, with two-hot targets and expectation decoding
(Section~\ref{sec:design}). To see what the grid costs, we decoded the same model's outputs four more ways
(Table~\ref{tab:decoding}).

\begin{table}[H]
\centering
\caption{Output resolution: the default timestamp model's logits decoded five ways. The model trains on two-hot targets over 20 ms timestamp tokens. Mean over the seven test sets of the mean absolute error, the median absolute error (both ms) and the shares within 10 and 20 ms (\%); TIMIT and Switchboard alone as MAE / within 20 ms.}
\label{tab:decoding}
\scriptsize
\setlength{\tabcolsep}{4pt}
\begin{adjustbox}{max width=\linewidth}
\begin{tabular}{lrrrrrr}
\toprule
\schead \scth{decoding} & \scth{MAE} & \scth{median} & \scth{10 ms} & \scth{20 ms} & \scth{TIMIT} & \scth{Switchboard} \\
\midrule
arg-max bin (the 20 ms grid) & 15.5 & 8.4 & 58.4 & 82.5 & 12.2/87.4 & 24.1/77.9 \\
expectation within 1 bin & 14.1 & 6.3 & 66.7 & 84.7 & 10.6/89.0 & 23.0/80.6 \\
expectation within 2 bins (default) & 14.1 & 6.5 & 66.1 & 84.8 & 10.5/89.1 & 23.0/80.7 \\
expectation within 5 bins & 14.1 & 6.8 & 64.2 & 83.7 & 10.6/88.0 & 23.2/79.1 \\
expectation over all bins & 17.0 & 9.5 & 52.6 & 76.1 & 15.8/74.7 & 25.6/72.7 \\
\bottomrule
\end{tabular}
\end{adjustbox}
\end{table}

Snapping to the most likely token, the grid itself, costs \val{s4/dec/grid-cost/mae}~ms of mean error,
\val{s4/dec/grid-cost/median}~ms of median error and \val{s4/dec/grid-cost/w10} points within 10~ms. The expectation
within one or two tokens recovers all of it: the position inside a frame is in the distribution. A wider window
loses a little, and the expectation over all tokens loses a lot (\val{s4/dec/full/mae}~ms), because a second, distant
mode drags it. The grid is not what limits the model's decoding.

\subsection{Finer timestamp tokens}
\label{sec:finer}

Does a finer timestamp vocabulary make the aligner more precise? The question follows from how well the
encoder-decoder design works. Predicting each boundary as one of Whisper's timestamp tokens, the timestamp model places
the most boundaries within 20~ms on every test set (Section~\ref{sec:families}), and on FA-Bench's TIMIT test it
cuts the best published error by \val{fb/ts/cut-pct}\% (Section~\ref{sec:sota}). Yet those tokens are 20~ms apart,
a wide step next to the slot-after recipe's median error of \val{res10/post/median}~ms over the seven test sets. If
the vocabulary holds the model back, finer tokens should help.

We test this by adding tokens. Whisper has one timestamp token every 20~ms, 1{,}501 in all. We add 1{,}500 new ones,
each halfway between two of them, so that there is a token every 10~ms. Each new token's embedding starts as the mean
of its two neighbours and trains beside the LoRA adapters. In our format the timestamps never enter the decoder's
input. The slots are special tokens, and only the classifier their states are scored against holds the timestamp
embeddings, so the new tokens need no other change to the model. We train the slot-after recipe with two-hot targets
at 10~ms and decode over the same window of $\pm 40$~ms (Table~\ref{tab:res10}).

\begin{table}[H]
\centering
\caption{Timestamp resolution, with the slots after the word. Whisper's own timestamp tokens are 20~ms apart; \emph{10~ms} adds one new token halfway between every two of them, \emph{5~ms} three. Means over the seven test sets in their own condition, then over seeds: mean and median absolute error in ms, share of boundaries within 10 and 20~ms in \%. The first block is the baseline: the best recipe of Section~\ref{sec:release}, the slots after the word with 10\% transcript noise in training, which has the lowest seven-set error of any arm, beside the same recipe on 10~ms tokens. The second compares the token spacings on the slot-after recipe without noise, with five seeds each at 20 and 10~ms. The third changes one thing in that 10~ms recipe. The last puts the 10~ms tokens on other checkpoints, each beside its 20~ms runs.}
\label{tab:res10}
\small
\setlength{\tabcolsep}{4pt}
\begin{adjustbox}{max width=\linewidth}
\begin{tabular}{llrrrrr}
\toprule
\schead \scth{tokens} & \scth{setting} & \scth{seeds} & \scth{mean} & \scth{median} & \scth{$\le$10~ms} & \scth{$\le$20~ms} \\
\midrule
20 ms & best recipe (baseline) & 2 & 12.68 & 6.51 & 65.9 & 85.2 \\
10 ms & best recipe, 1,500 new tokens & 1 & 12.74 & 6.41 & 66.1 & 84.6 \\
\midrule
20 ms & Whisper's tokens & 5 & 13.02 & 6.56 & 65.8 & 85.0 \\
10 ms & 1,500 new tokens & 5 & 12.99 & 6.47 & 66.0 & 84.8 \\
5 ms & 4,500 new tokens & 1 & 13.14 & 6.47 & 65.6 & 84.4 \\
\midrule
10 ms & new tokens initialised at random & 1 & 13.03 & 6.33 & 66.5 & 84.9 \\
10 ms & new tokens at lr $5{\times}10^{-3}$ & 1 & 12.87 & 6.50 & 66.0 & 84.8 \\
10 ms & new tokens at lr $2{\times}10^{-4}$ & 1 & 12.94 & 6.36 & 66.3 & 84.8 \\
10 ms & one-hot targets & 1 & 13.08 & 6.57 & 65.2 & 84.4 \\
10 ms & 4,000 steps & 1 & 13.04 & 6.50 & 66.1 & 84.8 \\
10 ms & LoRA rank 64 & 1 & 12.86 & 6.33 & 66.4 & 85.0 \\
\midrule
20 ms & vanilla large-v3 & 2 & 12.88 & 6.48 & 66.2 & 85.2 \\
10 ms & vanilla large-v3 & 1 & 12.71 & 6.36 & 66.5 & 85.2 \\
20 ms & large-v3-turbo & 2 & 12.99 & 6.54 & 65.9 & 84.9 \\
10 ms & large-v3-turbo & 1 & 12.93 & 6.54 & 65.9 & 84.6 \\
\bottomrule
\end{tabular}
\end{adjustbox}
\end{table}

\textbf{The finer tokens do not lower the error.} Over the seven test sets, \val{res10/ms10/n} seeds with the new tokens
score \val{res10/ms10/mae}~ms and \val{res10/post/n} seeds with Whisper's own score \val{res10/post/mae}~ms, a difference
of \val{res10/d-mae}~ms with a standard error of \val{res10/se-mae}. The best recipe, with transcript noise in
training, gives the same answer: \val{res10/best10/mae}~ms with the new tokens against \val{res10/best/mae} without
them (\val{res10/best10/n} seed against \val{res10/best/n}). Some boundaries do move closer. The median error goes
from \val{res10/post/median} to \val{res10/ms10/median}~ms, and the share within 10~ms from \val{res10/post/w10} to
\val{res10/ms10/w10}\%. But the share within 20~ms falls from \val{res10/post/w20} to \val{res10/ms10/w20}\%, and the
20 to 28~s TIMIT windows lose \val{res10/long/d}~ms (standard error \val{res10/long/se}). No test set gains by more
than its seed spread; the largest gain is on Switchboard (\val{res10/swb/d}~ms, standard error \val{res10/swb/se}). The
decoding experiment of Section~\ref{sec:resolution} explains why: the expectation already places a boundary inside its
20~ms step, so a finer grid has little left to recover. The mean error is twice the median, and the larger misses that set it are not a matter of
resolution.

\textbf{No setting of the new tokens changes the answer.} We then varied the 10~ms recipe one thing at a time, with one
seed each (Table~\ref{tab:res10}). Starting the new embeddings at random instead of between their neighbours scores
\val{res10/sw/rand/mae}~ms. Their own learning rate, $5{\times}10^{-3}$ or $2{\times}10^{-4}$ instead of $10^{-3}$,
scores \val{res10/sw/lr5e3/mae} or \val{res10/sw/lr2e4/mae}~ms. Twice the training scores \val{res10/sw/4k/mae}~ms, and
adapters of rank 64 score \val{res10/sw/r64/mae}~ms. One-hot targets on the nearest token score
\val{res10/sw/round/mae}~ms, with the fewest boundaries within 10~ms (\val{res10/sw/round/w10}\%), and 5~ms tokens,
three new ones between every two of Whisper's, score \val{res10/sw/ms5/mae}~ms. Every one of these runs falls inside
the range of the five seeds on Whisper's own tokens, \val{res10/post/lo} to \val{res10/post/hi}~ms. On the other
checkpoints the 10~ms tokens score \val{res10/sw/v3-10/mae}~ms against \val{res10/sw/v3-20/mae} on vanilla large-v3,
and \val{res10/sw/turbo-10/mae} against \val{res10/sw/turbo-20/mae} on large-v3-turbo (one run against two seeds
each), differences no larger than the spread between seeds. Whisper's 20~ms tokens, read as an expectation, are fine
enough.

\begin{figure}[H]
\centering
\includegraphics[width=0.88\linewidth]{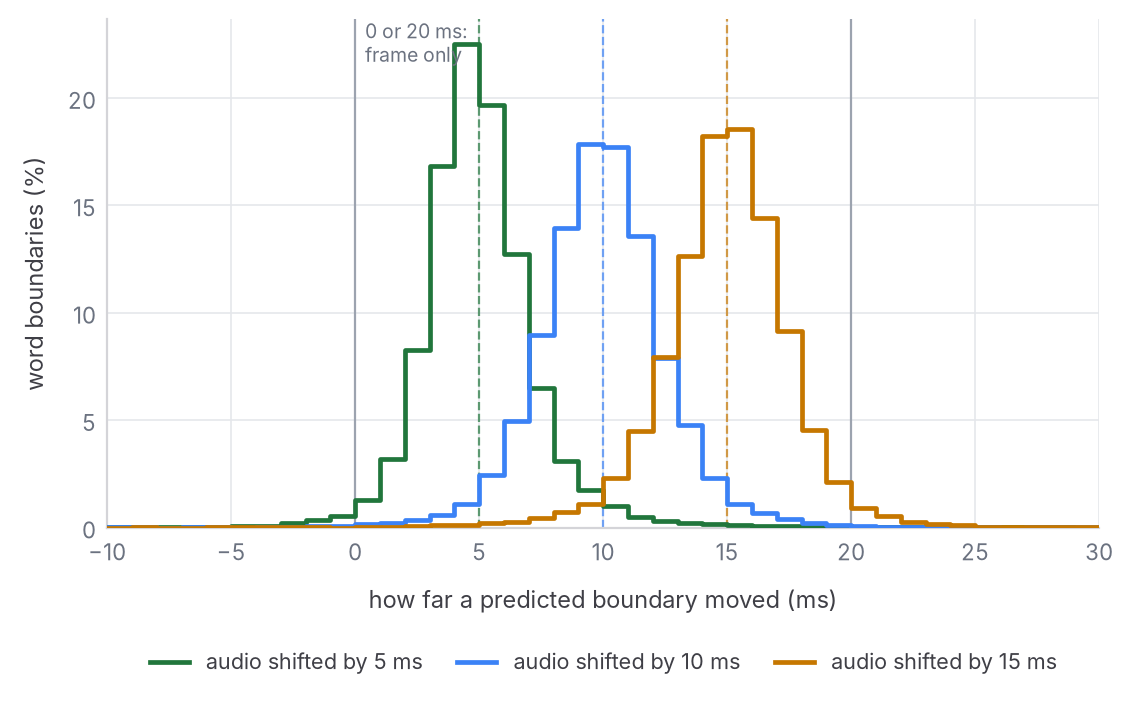}
\caption{How far each predicted boundary moves when a TIMIT test clip is shifted by part of Whisper's 20~ms encoder
frame, for the slot-after model on Whisper's 20~ms tokens (\val{sub/n} boundaries). A model that knew only which frame
a boundary falls in would move each boundary by 0 or 20~ms (grey lines). This one moves it by the shift (dashed
lines).}
\label{fig:subframe}
\end{figure}

\textbf{Is the encoder's frame the limit?} Whisper's encoder emits one frame every 20~ms, and the decoder sees nothing
finer, so a 20~ms floor is a natural suspicion. We tested it by shifting each TIMIT test clip by part of a frame,
prepending 5, 10 or 15~ms of silence, and measuring how far each predicted boundary moved. A model that knew only
which frame a boundary falls in would move each one by 0 or 20~ms. The 20~ms model moves them by the shift itself
(Figure~\ref{fig:subframe}). Shifted by half a frame, \val{sub/post/10/near}\% of its boundaries move by 7.5 to
12.5~ms and \val{sub/post/10/frame}\% by about 0 or 20~ms, and its error stays where it was (\val{sub/post/10/mae}
against \val{sub/post/0/mae}~ms). So the frames carry the position inside them, and the model reads it. The 10~ms
tokens read the same information (\val{sub/ms10/10/near}\% and \val{sub/ms10/10/frame}\%), a little less steadily:
shifted by one whole frame, \val{sub/ms10/20/near}\% of their boundaries move by 17.5 to 22.5~ms, against
\val{sub/post/20/near}\% for the 20~ms model.

\subsection{Slot order}
\label{sec:slots}

The decoder reads the transcript with two timestamp slots per word (Section~\ref{sec:design}). The slots can sit in
two places. \emph{Around} the word, the start slot comes before the word and the end slot after it:
\begin{center}
\footnotesize
\texttt{<|startoflm|> his <|startofprev|>}
\end{center}
This is Whisper's own segment order, and our default. \emph{After} the word, both slots follow it:
\begin{center}
\footnotesize
\texttt{his <|startoflm|> <|startofprev|>}
\end{center}
Whisper's decoder is causal. Each position sees the audio and the tokens before it, and nothing after. Around the word,
the start slot has seen the words before it, but not the word it is timing. After the word, it has seen that word
too. Table~\ref{tab:format} compares the two.

\begin{table}[H]
\centering
\caption{Where the timestamp slots sit, and Whisper's own format. Mean absolute error in ms of word boundaries; \emph{mean}: the seven test sets in their own condition, with the share within 20 ms; per set: word starts / word ends. Arms are means over their seeds; the four layouts differ in nothing else.}
\label{tab:format}
\scriptsize
\setlength{\tabcolsep}{3pt}
\begin{adjustbox}{max width=\linewidth}
\begin{tabular}{lrrrrrrrr}
\toprule
\schead \scth{decoder input} & \scth{seeds} & \scth{mean} & \scth{TIMIT} & \scth{Buckeye} & \scth{SWB} & \scth{Seoul} & \scth{IFA} & \scth{VoxAng.} \\
\midrule
slots around the word (Whisper's order) & 5 & 14.00/84.9 & 10.7/10.3 & 13.1/11.0 & 29.5/16.9 & 12.6/10.3 & 14.1/16.3 & 9.3/21.4 \\
slots after the word & 5 & 13.02/85.0 & 10.8/10.3 & 11.5/11.1 & 20.3/16.5 & 10.8/10.0 & 13.7/16.0 & 9.4/21.3 \\
around the word, rounded targets & 2 & 14.26/84.5 & 11.2/10.6 & 13.1/11.2 & 29.7/17.4 & 12.8/10.6 & 14.4/16.3 & 9.4/21.5 \\
Whisper's own layout & 2 & 19.12/83.2 & 12.9/10.7 & 27.4/24.0 & 36.8/26.0 & 24.3/21.2 & 14.6/16.7 & 9.5/20.7 \\
\bottomrule
\end{tabular}
\end{adjustbox}
\end{table}

\begin{figure}[H]
\centering
\includegraphics[width=0.88\linewidth]{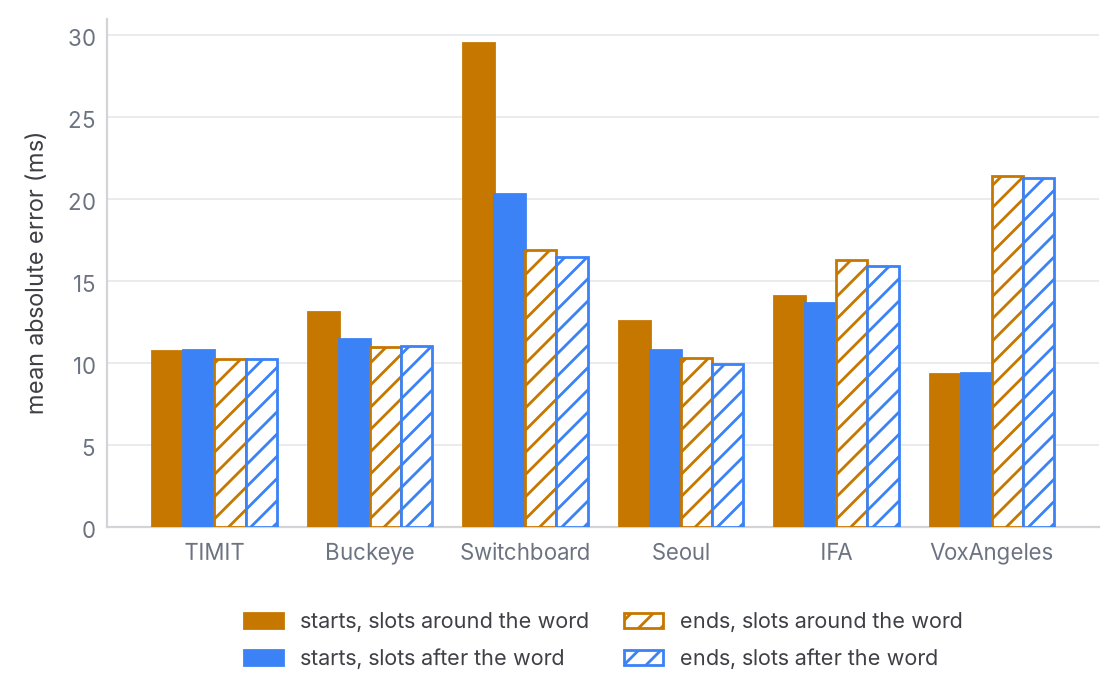}
\caption{Word-start and word-end error on each test set with the timestamp slots around the word (Whisper's order,
the start slot before the word) and both after it. Means over each layout's seeds.}
\label{fig:slots}
\end{figure}

\textbf{Slots after the word cut the error by \val{fmt/post/d-mae-abs}~ms}, from \val{fmt/pre/mean-mae} to
\val{fmt/post/mean-mae}~ms over the seven test sets, with \val{fmt/pre/runs} and \val{fmt/post/runs} seeds (standard
error of the difference \val{fmt/post/se-mae}~ms). That is \val{fmt/post/z-mae} standard deviations of the default's
seeds, more than any adapter setting moved it. The gain is all in the word starts, and all in
conversation (Figure~\ref{fig:slots}). On Switchboard's telephone conversations word starts go from
\val{fmt/pre/swb/start} to \val{fmt/post/swb/start}~ms, \val{fmt/post/swb/start-gain-pct}\% less, and the set's error
from \val{fmt/pre/swb/mae} to \val{fmt/post/swb/mae}~ms, \val{fmt/post/swb/gain-pct}\% less, while its word ends stay
at \val{fmt/post/swb/end}~ms. Buckeye's and Seoul's starts improve as well (\val{fmt/pre/buckeye/start} to
\val{fmt/post/buckeye/start} and \val{fmt/pre/seoul/start} to \val{fmt/post/seoul/start}~ms); TIMIT's read sentences
do not move (\val{fmt/pre/timit/start} and \val{fmt/post/timit/start}~ms). In conversation the next word is hard to
guess; in read speech an onset is easy to find whatever the word.

\textbf{Whisper's own layout is worse.} Trained and decoded the way Whisper handles timestamps, with real timestamp
tokens rounded to 20~ms in the input, next-token cross-entropy over the whole vocabulary, and the slots filled in
order, each with the most likely timestamp no earlier than the one before, the model scores
\val{fmt/whisper/mean-mae}~ms, \val{fmt/whisper/d-mae}~ms against the default. A slot that jumps ahead drags every
word after it, which hits long conversational windows hardest (Buckeye \val{fmt/whisper/buckeye/mae}~ms, Seoul
\val{fmt/whisper/seoul/mae}), and filling the slots one at a time is several times slower than reading them in one
pass. Rounding the targets alone, in the default layout, costs \val{fmt/round/d-mae}~ms: the two-hot targets are
worth keeping.

\subsection{Phone edges}

Most of the training corpora mark phones as well as words. To see whether phone edges sharpen word edges, a linear
head on the encoder output learns, during training only, to mark word edges per 20~ms frame, or word and phone edges
where the corpus has phones (TIMIT, Seoul, IFA and VoxAngeles), with two seeds each. The head is dropped at inference,
so the aligner itself is unchanged (Table~\ref{tab:aux}).

\begin{table}[H]
\centering
\caption{Boundary supervision: the default timestamp model, and the same model trained with a linear head on the encoder output that marks word edges, or word and phone edges, per 20 ms frame (dropped at inference). Mean absolute error in ms / share within 20 ms in \%, arm means over seeds; \emph{mean} over the seven test sets in their own condition.}
\label{tab:aux}
\scriptsize
\setlength{\tabcolsep}{3.5pt}
\begin{adjustbox}{max width=\linewidth}
\begin{tabular}{lrrrrrrrr}
\toprule
\schead \scth{extra targets} & \scth{seeds} & \scth{TIMIT} & \scth{Buckeye} & \scth{SWB} & \scth{Seoul} & \scth{IFA} & \scth{VoxAng.} & \scth{mean} \\
\midrule
none (default) & 5 & 10.5/89.1 & 12.0/86.1 & 23.2/80.8 & 11.4/89.9 & 15.2/78.5 & 15.4/80.5 & 14.00/84.91 \\
word edges & 2 & 10.5/89.2 & 12.2/86.2 & 24.2/80.4 & 11.3/90.1 & 14.8/78.6 & 15.7/79.6 & 14.15/84.79 \\
word and phone edges & 2 & 10.5/89.4 & 12.1/86.0 & 23.4/80.6 & 11.3/90.2 & 15.0/78.6 & 15.5/80.2 & 14.01/84.96 \\
\bottomrule
\end{tabular}
\end{adjustbox}
\end{table}

Neither target moves word boundaries. Word edges change the mean error over the seven sets by
\val{s4/aux/word/d-mae}~ms (standard error \val{s4/aux/word/se-mae}), word and phone edges by
\val{s4/aux/wordphone/d-mae}~ms (\val{s4/aux/wordphone/se-mae}). The timestamp model already learns where words
begin and end from its own targets, and a frame-level loss on the encoder adds nothing to it.

\subsection{Telephone augmentation}

Every run so far sent half of its training clips through a telephone condition. Stage 5 varies that share, $p$,
from none to all, with three seeds at none and at a quarter and \val{s3/n-seeds} at the default half
(Table~\ref{tab:augmentation}). Two codecs never
enter training: AMR-NB, which mobile networks use, and G.723.1, a VoIP codec. Their delays, \val{timit/lag/amrnb} and
\val{timit/lag/g723}~ms, are measured and removed like the others'.

\begin{table}[H]
\centering
\caption{Telephone augmentation (stage 5): $p$ is the share of training clips sent through one of the six conditions of Table~\ref{tab:channel}, with white noise on half of them. AMR-NB (12.2 kbit/s) and G.723.1 (6.3 kbit/s) never enter training. Mean absolute error in ms / share within 20 ms in \%; \emph{clean} averages the six test sets in their clean condition, each codec column its codecs. Arms with several seeds show their mean. Bold: the lowest error per column (ties at the shown precision all bold).}
\label{tab:augmentation}
\scriptsize
\setlength{\tabcolsep}{3.5pt}
\begin{adjustbox}{max width=\linewidth}
\begin{tabular}{lrrrrrrr}
\toprule
\schead \scth{$p$} & \scth{runs} & \scth{clean} & \scth{TIMIT, six codecs} & \scth{TIMIT, unseen} & \scth{Buckeye, three codecs} & \scth{Buckeye, unseen} & \scth{Switchboard} \\
\midrule
0 & 3 & 12.5/85.4 & 17.1/74.6 & 17.6/73.8 & 15.4/81.3 & 15.8/80.4 & 24.5/80.0 \\
0.25 & 3 & \textbf{12.4}/85.6 & 11.2/87.8 & 12.3/86.3 & 13.0/84.8 & 13.3/84.1 & 23.3/80.5 \\
0.5 (default) & 5 & 12.5/85.6 & 11.2/88.0 & 12.2/86.4 & \textbf{12.9}/84.9 & \textbf{13.2}/84.3 & 23.2/80.8 \\
0.75 & 1 & 12.6/85.1 & \textbf{11.0}/88.1 & \textbf{12.1}/86.8 & 13.1/84.3 & 13.3/83.9 & \textbf{23.0}/80.7 \\
1 & 1 & 13.0/84.8 & 11.1/88.2 & \textbf{12.1}/86.9 & 13.1/84.6 & 13.4/83.9 & \textbf{23.0}/80.5 \\
\midrule
MFA (bar) & & -- & 25.7/62.7 & 25.4/63.1 & 26.6/64.5 & 26.0/64.9 & 36.5/64.2 \\
\bottomrule
\end{tabular}
\end{adjustbox}
\end{table}

\textbf{Without augmentation the channel costs \val{s5/p0/timit/cost-pct}\%.} Trained on clean audio alone, the
model scores \val{s5/p0/timit/clean}~ms on clean TIMIT and \val{s5/p0/timit/codec-lo} to
\val{s5/p0/timit/codec-hi}~ms through the six codecs. Most of that is the narrow band, not the compression:
band-limiting alone costs it \val{s5/p0/timit/pcm-cost}~ms. With the default augmentation the six codecs cost
\val{s5/p05/timit/cost-lo} to \val{s5/p05/timit/cost-hi}~ms. Over the codecs the augmentation uses, the default is
\val{s5/diff/seen/mae-abs}~ms better than no augmentation (standard error \val{s5/diff/seen/mae-se}) and places \val{s5/diff/seen/w20-abs} points more boundaries within 20~ms.

\textbf{The gain carries over to codecs it never used.} On AMR-NB and G.723.1 the default is
\val{s5/diff/unseen/mae-abs}~ms better (standard error \val{s5/diff/unseen/mae-se}) and \val{s5/diff/unseen/w20-abs}
points higher within 20~ms, nearly the gain on the codecs it trained with. AMR-NB keeps the largest residual cost:
\val{s5/p05/timit/amrnb}~ms on TIMIT against \val{s5/p05/timit/clean} clean. Adding AMR-NB to the augmentation
hardly moves it (\val{s5/augamr/timit/amrnb}~ms over two seeds, against
\val{s5/p05/timit/amrnb-lo} to \val{s5/p05/timit/amrnb-hi} for the default's), so what remains is the codec's own loss
of information, not unfamiliarity.

\textbf{Real telephone lines gain less, and clean speech pays nothing.} Switchboard improves by
\val{s5/diff/swb/mae-abs}~ms (standard error \val{s5/diff/swb/mae-se}). On clean speech the default and no
augmentation differ by \val{s5/diff/clean/mae}~ms (standard error \val{s5/diff/clean/mae-se}), and up to $p = 0.75$
nothing changes. Only degrading every clip costs clean speech: \val{s5/p1/timit-clean/mae}~ms on clean TIMIT against
\val{s5/p05/timit-clean/mae}, with no further telephone gain. A quarter of the clips already earns the full gain, and
the default stays at a half.

\subsection{Call-centre augmentation}
\label{sec:realcalls}

The six conditions are generic: a band-pass from 300 to 3{,}400~Hz, a codec, and line noise on half the clips. For
the release we also used the telephone degradation pipeline we built for CallEnhancer, our open-source call-centre
speech restoration model\footnote{\url{https://huggingface.co/Scicom-intl/CallEnhancer}}, and degraded every
training clip once, offline. We compare its output with \val{tel/real/n-legs} legs of \val{tel/real/n-calls} real
call-centre recordings, two-channel 8~kHz calls of which only aggregate statistics leave the company.

The degrader does not keep time. A median clip comes back \val{tel/lag/median}~ms late, but \val{tel/lag/over5-pct}\%
come back more than 5~ms late and the latest \val{tel/lag/max}~ms. So every degraded clip is cross-correlated with its clean input between 300 and 3{,}000~Hz, shifted back and cut to
its length; afterwards all but \val{tel/lag/residual-n} of the \val{tel/lag/kept} clips line up to the sample. Where
the correlation peak stays below 0.3, the degradation has swamped the speech and its lag cannot be trusted: those
\val{tel/lag/dropped-pct}\% of clips train clean.

\begin{figure}[H]
\centering
\includegraphics[width=0.9\linewidth]{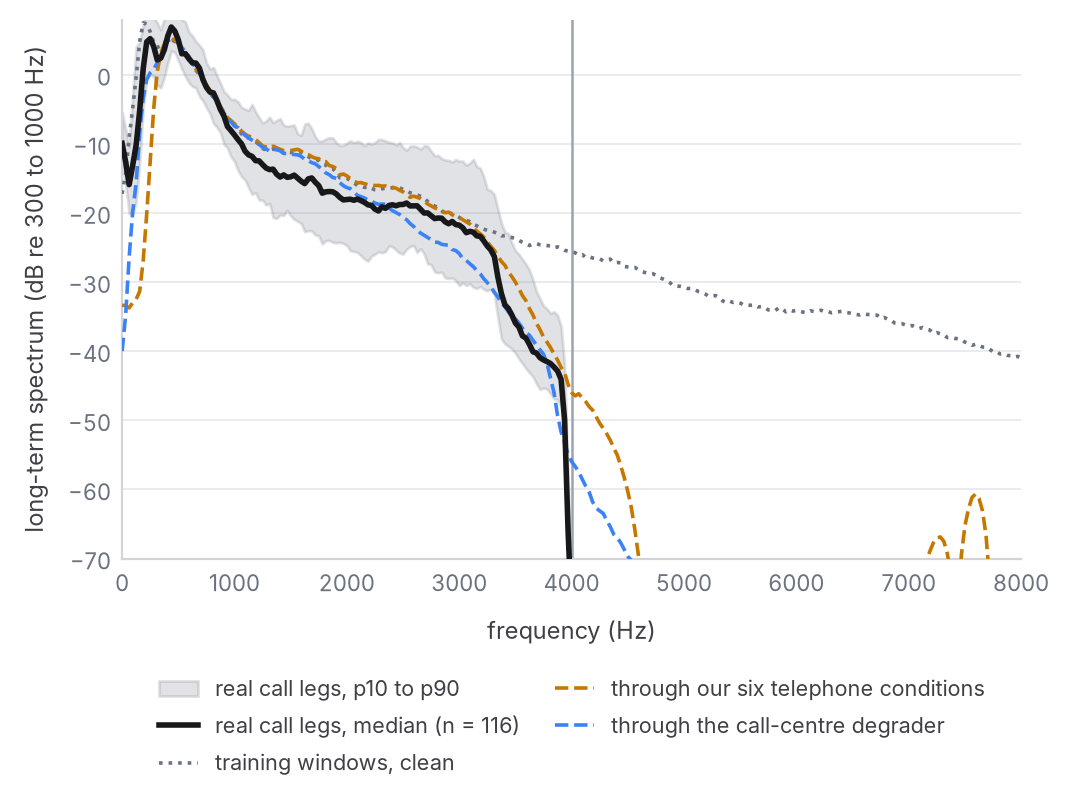}
\caption{Long-term spectrum of real call legs (median and 10th to 90th percentile band over \val{tel/real/n-legs}
legs) and of our training windows, clean, through our six telephone conditions and through the call-centre
degrader (medians over windows), each normalised to its own 300 to 1000~Hz level. The vertical line marks 4~kHz, the
Nyquist limit of the real 8~kHz recordings.}
\label{fig:telltas}
\end{figure}

We measured \val{tel/n-windows} of our training windows with the same statistics as the real legs
(Table~\ref{tab:telinput}),
clean, through the six conditions and through the call-centre degrader. Between 300 and 3{,}400~Hz every input has the
legs' spectral shape, a mean \val{tel/ltas-dev-lo} to \val{tel/ltas-dev-hi}~dB from their median; the edges are what
differ (Figure~\ref{fig:telltas}). The degrader's long-term spectrum stays inside the legs' 10th to 90th percentile
band from \val{tel/sidon/inband-lo} to \val{tel/sidon/inband-hi}~Hz, and its tilt, \val{tel/sidon/tilt}~dB per
octave, matches theirs, \val{tel/real/tilt}. The six conditions leave the band below \val{tel/codecs/inband-lo}~Hz,
because their band-pass removes everything under 300~Hz: their low edge is \val{tel/codecs/lf}~Hz, against the legs'
\val{tel/real/lf} and the degrader's \val{tel/sidon/lf}. The degrader's high edge is a little low,
\val{tel/sidon/edge} against \val{tel/real/edge}~Hz.

\begin{table}[H]
\centering
\caption{Our training windows against real call legs, medians: 1{,}000 training windows clean, through our six telephone conditions as training draws them, and through the call-centre degrader (959 of them: the rest failed the lag check), against 116 legs of 58 real two-channel 8~kHz call-centre recordings. Speech: share of 20~ms frames within 25~dB of the 90th-percentile frame; SNR: 90th over 10th percentile of frame energy; floor: the 10th percentile; level: mean energy of the speech frames; edges: lowest and highest frequency of the long-term spectrum within 15 and 25~dB of its 300 to 1000~Hz level; tilt: its slope over 0.5 to 2.5~kHz; above 4~kHz: share of windows whose high edge lies there.}
\label{tab:telinput}
\scriptsize
\setlength{\tabcolsep}{3pt}
\begin{adjustbox}{max width=\linewidth}
\begin{tabular}{lrrrrrrrrr}
\toprule
\schead \scth{input} & \shortstack{\scth{length}\\\scth{(s)}} & \shortstack{\scth{speech}\\\scth{share}} & \shortstack{\scth{SNR}\\\scth{(dB)}} & \shortstack{\scth{floor}\\\scth{(dBFS)}} & \shortstack{\scth{level}\\\scth{(dBFS)}} & \shortstack{\scth{low edge}\\\scth{(Hz)}} & \shortstack{\scth{high edge}\\\scth{(Hz)}} & \shortstack{\scth{tilt}\\\scth{(dB/oct)}} & \shortstack{\scth{above}\\\scth{4~kHz}} \\
\midrule
real call legs & 29.9 & 38\% & 50.9 & $-$81 & $-$26 & 94 & 3{,}281 & $-$10.7 & 0\% \\
\midrule
our windows, clean & 2.5 & 82\% & 30.5 & $-$58 & $-$30 & 62 & 4{,}578 & $-$9.1 & 64\% \\
\quad six conditions & 2.5 & 77\% & 31.5 & $-$63 & $-$33 & 250 & 3{,}312 & $-$9.1 & 3\% \\
\quad call-centre degrader & 2.5 & 78\% & 33.2 & $-$50 & $-$19 & 156 & 3{,}000 & $-$10.4 & 8\% \\
\bottomrule
\end{tabular}
\end{adjustbox}
\end{table}

\begin{figure}[H]
\centering
\includegraphics[width=0.78\linewidth]{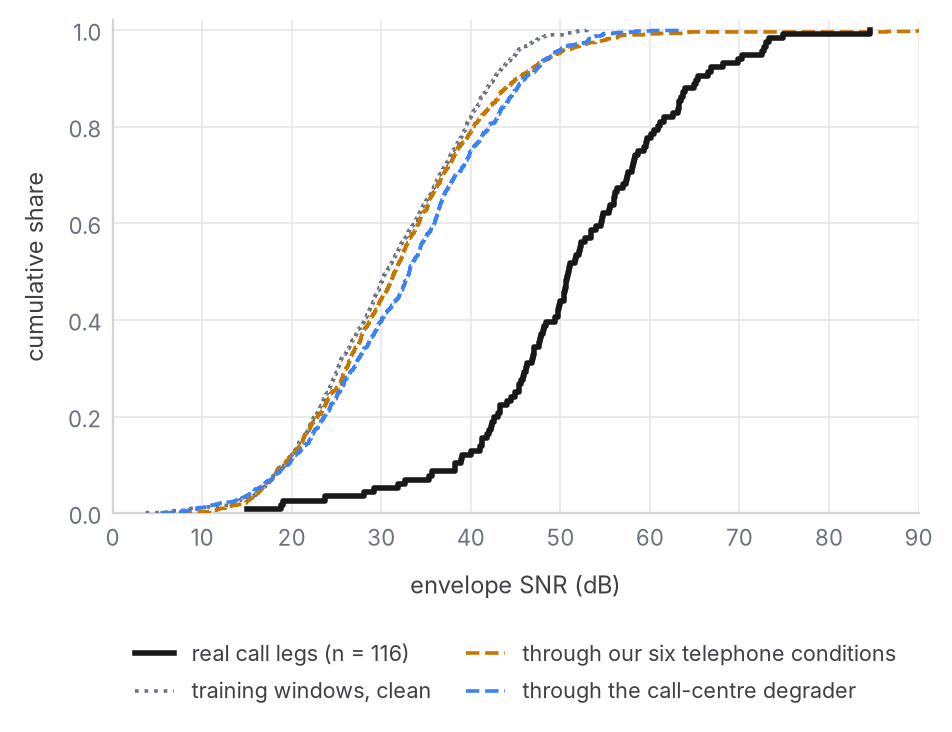}
\caption{Envelope SNR (90th over 10th percentile of 20~ms frame energy), cumulative over real legs and over our
training windows.}
\label{fig:telsnr}
\end{figure}

The noise statistics cannot match, and the clean windows show why (Figure~\ref{fig:telsnr}). A real leg lasts
\val{tel/real/dur}~s at the median, and only \val{tel/real/speech}\% of it is that party's speech. The rest is the
other party's turn at the line's floor, so the 10th percentile of its frame energy reads the line, at
\val{tel/real/floor}~dBFS, and its envelope SNR is \val{tel/real/snr}~dB. Our windows last \val{tel/clean/dur}~s and
are \val{tel/clean/speech}\% speech, so the same percentile falls on speech, and the clean windows themselves read
\val{tel/clean/snr}~dB. On that scale neither augmentation drowns the speech: the six conditions read
\val{tel/codecs/snr}~dB and the call-centre degrader \val{tel/sidon/snr}~dB.

\begin{figure}[H]
\centering
\includegraphics[width=0.78\linewidth]{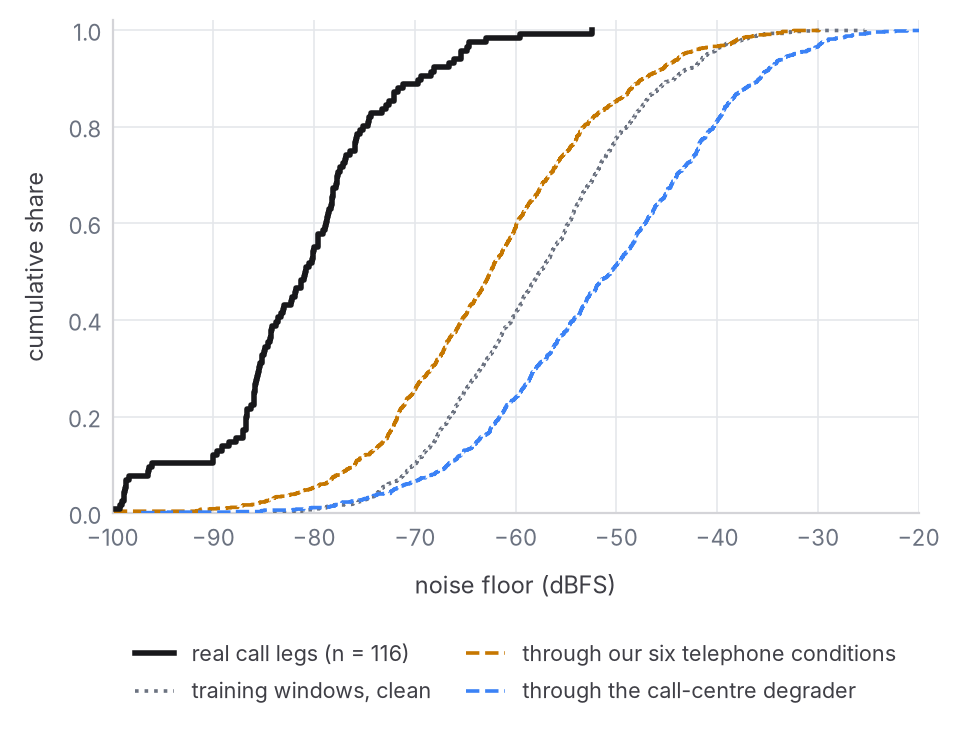}
\caption{Noise floor (10th percentile of 20~ms frame energy), cumulative over real legs and over our training
windows.}
\label{fig:telfloor}
\end{figure}

The floor tells the same story from the other end (Figure~\ref{fig:telfloor}): the degrader raises the windows' floor
by \val{tel/sidon/floor-rise}~dB and their speech level by \val{tel/sidon/level-rise}~dB, so it adds the channel of a
real call, not its silences.

\textbf{Test sets through the degrader.} The codec conditions of Section~\ref{sec:channel} carry only a band limit and
a codec. To see what a fuller call costs, we put TIMIT test and Buckeye's held-out speakers through the same
degrader, one draw per clip, with its delay removed the same way (Table~\ref{tab:ce}). It is the released models' own
training augmentation, so this test favours them, as the codec conditions favour the models trained on those codecs.
The released large-v3 model goes from \val{ce/rel-v3/timit/clean/mae}~ms on clean TIMIT to
\val{ce/rel-v3/timit/ce/mae}~ms through the degrader, and keeps \val{ce/rel-v3/timit/ce/w20}\% of its boundaries
within 20~ms. Trained on the six conditions instead, the default model goes from \val{ce/default/timit/clean/mae} to
\val{ce/default/timit/ce/mae}~ms. Without telephone augmentation the error reaches \val{ce/aug0/timit/ce/mae}~ms.
MFA's more than doubles, from \val{ce/mfa/timit/clean/mae} to \val{ce/mfa/timit/ce/mae}~ms, and it fails on
\val{ce/mfa/timit/ce/failed} utterances. Buckeye's conversations keep the same order.

\begin{table}[H]
\centering
\caption{Test sets through the call-centre degrader, the release recipe's own training augmentation (one draw per clip, delay removed), against the clean audio and the codec conditions (the mean over TIMIT's six and Buckeye's three). Mean absolute error in ms / share within 20~ms in \%, on each condition's common utterances; Buckeye is its held-out speakers s01 to s04. The degraded condition favours the released models as the codec conditions favour the models trained on those codecs. Means over each recipe's seeds.}
\label{tab:ce}
\scriptsize
\setlength{\tabcolsep}{3pt}
\begin{adjustbox}{max width=\linewidth}
\begin{tabular}{lrrrrrr}
\toprule
\schead & \multicolumn{3}{c}{\scth{TIMIT test}} & \multicolumn{3}{c}{\scth{Buckeye}} \\
\cmidrule(lr){2-4}\cmidrule(lr){5-7}
\schead \scth{aligner} & \scth{clean} & \scth{codecs} & \scth{degrader} & \scth{clean} & \scth{codecs} & \scth{degrader} \\
\midrule
released, large-v3 & 10.6 / 89 & 11.8 / 86 & 13.4 / 83 & 11.2 / 86 & 12.5 / 85 & 13.5 / 82 \\
released, large-v3-turbo & 10.9 / 88 & 11.8 / 86 & 14.1 / 82 & 11.3 / 86 & 12.7 / 84 & 14.5 / 81 \\
slots after + transcript noise & 10.7 / 89 & 11.3 / 87 & 16.4 / 78 & 11.1 / 86 & 11.7 / 86 & 15.9 / 80 \\
slots after the word, six conditions & 10.5 / 89 & 11.2 / 88 & 16.3 / 78 & 11.3 / 86 & 12.0 / 86 & 15.9 / 80 \\
default, six conditions & 10.5 / 89 & 11.2 / 88 & 16.6 / 79 & 12.0 / 86 & 12.9 / 85 & 17.0 / 79 \\
no telephone augmentation & 10.4 / 89 & 17.1 / 75 & 20.9 / 70 & 12.3 / 86 & 15.4 / 81 & 18.6 / 76 \\
CTC, Whisper-Scicom encoder & 16.4 / 76 & 16.4 / 76 & 19.9 / 70 & 14.0 / 81 & 14.7 / 79 & 17.4 / 75 \\
MFA, English ARPA & 19.1 / 68 & 25.7 / 63 & 43.1 / 53 & 21.5 / 71 & 26.6 / 65 & 46.1 / 55 \\
\bottomrule
\end{tabular}
\end{adjustbox}
\end{table}

The release recipe trains with this degrader in place of the six conditions, together with the slots after the word
and transcript noise. Two seeds of each step toward it, on a newer release of Whisper-Scicom fine-tuned on the same
private data and on the public Whisper checkpoints, are compared in Section~\ref{sec:release}.

\subsection{Whisper checkpoint}
\label{sec:checkpoint}

Every model so far was trained on Whisper-Scicom, our Whisper large-v3 fine-tuned on our private call-centre dataset. To
see whether that adaptation matters, and whether the 32-layer decoder does, we trained the default recipe on vanilla
Whisper large-v3 and on large-v3-turbo, which keeps the encoder and has 4 decoder layers, with \val{bb/v3/runs} and
\val{bb/turbo/runs} seeds (Table~\ref{tab:backbone}).

\begin{table}[H]
\centering
\caption{The checkpoint the adapters train on, with the default recipe: Whisper-Scicom (fine-tuned on Malaysian call-centre speech), vanilla Whisper large-v3, and large-v3-turbo (the same encoder, 4 decoder layers instead of 32). Mean absolute error in ms / share within 20 ms in \%, arm means over seeds; \emph{mean} over the seven test sets in their own condition. RTF: GPU seconds per second of audio on clean TIMIT, batch 16, on shared GPUs.}
\label{tab:backbone}
\scriptsize
\setlength{\tabcolsep}{3pt}
\begin{adjustbox}{max width=\linewidth}
\begin{tabular}{lrrrrrrrrrr}
\toprule
\schead \scth{checkpoint} & \scth{seeds} & \scth{TIMIT} & \scth{codecs} & \scth{Buckeye} & \scth{SWB} & \scth{Seoul} & \scth{IFA} & \scth{VoxAng.} & \scth{mean} & \scth{RTF} \\
\midrule
Whisper-Scicom & 5 & 10.5/89.1 & 11.2/88.0 & 12.0/86.1 & 23.2/80.8 & 11.4/89.9 & 15.2/78.5 & 15.4/80.5 & 14.00/84.91 & 0.0090 \\
Whisper large-v3 & 3 & 10.4/89.3 & 11.1/88.2 & 12.0/86.2 & 24.2/80.4 & 11.4/89.9 & 15.4/78.7 & 15.5/80.4 & 14.16/84.93 & 0.0094 \\
Whisper large-v3-turbo & 3 & 10.4/89.4 & 11.1/88.1 & 12.2/85.7 & 23.1/80.5 & 11.5/89.8 & 14.7/79.3 & 15.7/80.4 & 14.01/84.94 & 0.0079 \\
\bottomrule
\end{tabular}
\end{adjustbox}
\end{table}

\textbf{The call-centre fine-tune barely matters for alignment.} Vanilla large-v3 scores \val{bb/v3/mean-mae}~ms
against Whisper-Scicom's \val{bb/scicom/mean-mae} (a difference of \val{bb/v3/d-mae}~ms, standard error
\val{bb/v3/se-mae}), and through TIMIT's simulated codecs the two are level (\val{bb/v3/codecs/mae} against
\val{bb/scicom/codecs/mae}~ms). On Switchboard's real calls vanilla large-v3 is behind, \val{bb/v3/swb/mae} against
\val{bb/scicom/swb/mae}~ms (\val{bb/v3/swb/d}, standard error \val{bb/v3/swb/se}), which would credit the call-centre
fine-tune; but turbo, which has no call-centre fine-tune either, is not behind there (\val{bb/turbo/swb/mae}~ms), so the
evidence is weak.

\textbf{Turbo loses nothing on the test sets.} With an eighth of the decoder it scores \val{bb/turbo/mean-mae}~ms,
\val{bb/turbo/d-mae}~ms against Whisper-Scicom (standard error \val{bb/turbo/se-mae}, \val{bb/turbo/runs} seeds), and
aligns \val{bb/turbo/faster-pct}\% faster: the encoder, which both share, does most of the work. One sign points the other
way: on the development set turbo's mean error is \val{bb/turbo/dev-mae-lo} to \val{bb/turbo/dev-mae-hi}~ms against
\val{bb/scicom/dev-mae-lo} to \val{bb/scicom/dev-mae-hi} for Whisper-Scicom, while its median matches
(\val{bb/turbo/dev-median-lo} to \val{bb/turbo/dev-median-hi}~ms against \val{bb/scicom/dev-median-lo} to
\val{bb/scicom/dev-median-hi}): a few large errors that no test set repeats.

\subsection{Long recordings}

Calls last minutes, and the timestamp model reads one 30~s window, so a wrapper aligns any length in three steps.
MMS-300M, which reads a recording of any length, places every word roughly. The words are cut into runs of at most
25~s at the widest pauses it finds. Each run is aligned on its own crop of the audio, from the middle of the pause
before it to the middle of the pause after it, and its times are shifted back. A recording of 25~s or less goes to
the aligner whole, so no other number in this paper changes.

Only Seoul has held-out stretches long enough and fully labelled: \val{longseoul/n-items} stretches of 40 to
\val{longseoul/max-s}~s from its test speakers, \val{longseoul/n-words} words in \val{longseoul/hours}~h, with no
unlabelled speech inside. Buckeye's test recordings are segments of at most 25~s, and IFA's are short files.

\begin{table}[H]
\centering
\caption{Long recordings: the 61 continuously labelled stretches of 40 to 105 s of the held-out Seoul speakers (0.93 h), against the same speakers' windows of at most 30 s. Mean absolute error in ms / median in ms / share within 20 ms in \% / share beyond 500 ms in \%. The wrapper places words with MMS-300M, cuts them into runs of at most 25 s at pauses and aligns each run on its own crop. RTF: GPU seconds per second of audio on a shared GPU, coarse pass included.}
\label{tab:longform}
\scriptsize
\setlength{\tabcolsep}{4pt}
\begin{adjustbox}{max width=\linewidth}
\begin{tabular}{lrrr}
\toprule
\schead \scth{aligner} & \scth{long stretches} & \scth{30 s windows} & \scth{RTF} \\
\midrule
timestamps, LoRA, through the wrapper & 12.2 / 5.4 / 90.0 / 0.19 & 11.3 / 5.3 / 90.0 / 0.09 & 0.008 \\
\quad seed 1 & 11.6 / 5.4 / 90.0 / 0.15 & 11.5 / 5.3 / 89.9 / 0.10 & 0.007 \\
\quad seed 2 & 11.8 / 5.3 / 89.9 / 0.16 & 11.5 / 5.3 / 89.9 / 0.10 & 0.008 \\
CTC, Whisper-Scicom encoder, through the wrapper & 13.9 / 7.0 / 86.3 / 0.18 & 13.5 / 7.3 / 86.0 / 0.06 & 0.006 \\
CTC, MMS-300M fine-tuned, whole recording & 15.6 / 7.9 / 82.9 / 0.19 & 16.2 / 8.3 / 81.0 / 0.13 & 0.003 \\
MFA, \texttt{korean\_mfa} & 19.0 / 10.7 / 71.2 / 0.11 & 21.3 / 10.9 / 70.2 / 0.21 & 0.013 \\
MMS-300M alone (the coarse pass) & 51.6 / 26.6 / 38.0 / 0.64 & -- & 0.005 \\
\bottomrule
\end{tabular}
\end{adjustbox}
\end{table}

\textbf{Long recordings cost the timestamp model almost nothing} (Table~\ref{tab:longform}). Through the wrapper its
three seeds score \val{longseoul/ts-seeds/mae-lo} to \val{longseoul/ts-seeds/mae-hi}~ms, with
\val{longseoul/ts-seeds/w20-lo} to \val{longseoul/ts-seeds/w20-hi}\% of boundaries within 20~ms, as on the same
speakers' 30~s windows (\val{seoul/ft-ts-lora/clean/mae}~ms, \val{seoul/ft-ts-lora/clean/w20}\%). Gross errors rise
a little: boundaries off by more than half a second go from \val{seoul/ft-ts-lora/clean/over500}\% to
\val{longseoul/ft-ts-lora-long/clean/over500}\%. The refinement does the work, since the coarse pass alone is off by
\val{longseoul/cfa-segment/clean/mae}~ms on average. Every aligner keeps its place from the windowed test, and MFA,
which reads whole recordings, is no worse on them (\val{longseoul/mfa-korean/clean/mae} against
\val{seoul/mfa-korean/clean/mae}~ms). The pipeline runs at a real-time factor of
\val{longseoul/ft-ts-lora-long/clean/rtf} on one shared GPU, coarse pass included. One corpus, one language and four
speakers make this a first check, not a verdict.

\subsection{Transcript noise}
\label{sec:noise}

Outside a labelled corpus the transcript usually comes from a recogniser, not an annotator. We perturbed the
transcripts of TIMIT test and the Seoul held-out set: each word, with probability 10\%, is substituted by another word of the set, deleted, or
followed by an inserted word that is not spoken, and the three kinds are mixed at 20\%. Every aligner sees the same
draw, and we score the words the noise left unchanged, which shows how far an error disturbs its neighbours
(Table~\ref{tab:noise}).

\begin{table}[H]
\centering
\caption{Transcript noise: each word of the transcript substituted by another word of the set, deleted, or followed by an inserted word that is not spoken, with probability 10\% (20\% mixed: the three kinds equally), the same draw for every aligner. Scored on the words the noise left unchanged: mean absolute error in ms / share within 20 ms in \% / share beyond 500 ms in \%, on the utterances every aligner aligned in every copy. \emph{none}: the clean transcript, on the words the 10\% mixed copy leaves unchanged. \emph{Released}: the released large-v3 model (slots after the word, the call-centre degrader, 10\% noise in training). MFA: \texttt{english\_us\_arpa} on TIMIT, \texttt{korean\_mfa} on Seoul.}
\label{tab:noise}
\scriptsize
\setlength{\tabcolsep}{3pt}
\begin{adjustbox}{max width=\linewidth}
\begin{tabular}{llrrrrr}
\toprule
\schead \scth{set} & \scth{noise} & \scth{timestamps} & \scth{+10\% in training} & \scth{released, large-v3} & \scth{CTC} & \scth{MFA} \\
\midrule
TIMIT & none & 10.4/89.2/0.01 & 10.6/88.9/0.00 & 10.6/89.0/0.00 & 16.3/76.1/0.00 & 19.1/68.2/0.00 \\
 & 10\% substituted & 19.8/86.4/0.60 & 17.9/86.8/0.41 & 10.8/89.0/0.01 & 18.7/73.9/0.02 & 32.2/63.4/0.72 \\
 & 10\% deleted & 28.2/84.8/1.21 & 24.8/85.0/0.86 & 10.7/88.9/0.01 & 32.3/72.7/1.04 & 64.7/59.8/3.36 \\
 & 10\% inserted & 18.9/86.1/0.33 & 14.3/87.8/0.18 & 10.7/89.0/0.01 & 25.2/69.2/0.07 & 35.0/60.9/0.57 \\
 & 20\% mixed & 34.9/82.1/1.42 & 28.5/84.1/1.15 & 10.9/88.9/0.02 & 36.1/67.6/0.86 & 73.0/54.1/3.83 \\
\midrule
Seoul & none & 11.6/89.8/0.10 & 11.9/89.7/0.11 & 10.7/90.0/0.03 & 13.7/85.8/0.06 & 21.7/69.9/0.24 \\
 & 10\% substituted & 34.9/86.2/1.68 & 26.6/87.2/1.06 & 11.0/89.8/0.04 & 16.2/84.1/0.17 & 29.7/67.6/0.64 \\
 & 10\% deleted & 44.3/85.0/2.31 & 38.3/85.4/1.92 & 12.3/89.6/0.12 & 41.8/81.9/2.14 & 65.1/65.5/3.23 \\
 & 10\% inserted & 33.6/85.7/1.63 & 22.3/87.9/0.83 & 11.4/89.8/0.07 & 26.9/78.7/0.49 & 47.9/63.4/1.54 \\
 & 20\% mixed & 67.6/80.8/4.02 & 45.6/83.9/2.46 & 12.5/89.6/0.15 & 45.2/77.2/2.09 & 66.9/61.6/2.76 \\
\bottomrule
\end{tabular}
\end{adjustbox}
\end{table}

\begin{figure}[H]
\centering
\includegraphics[width=0.88\linewidth]{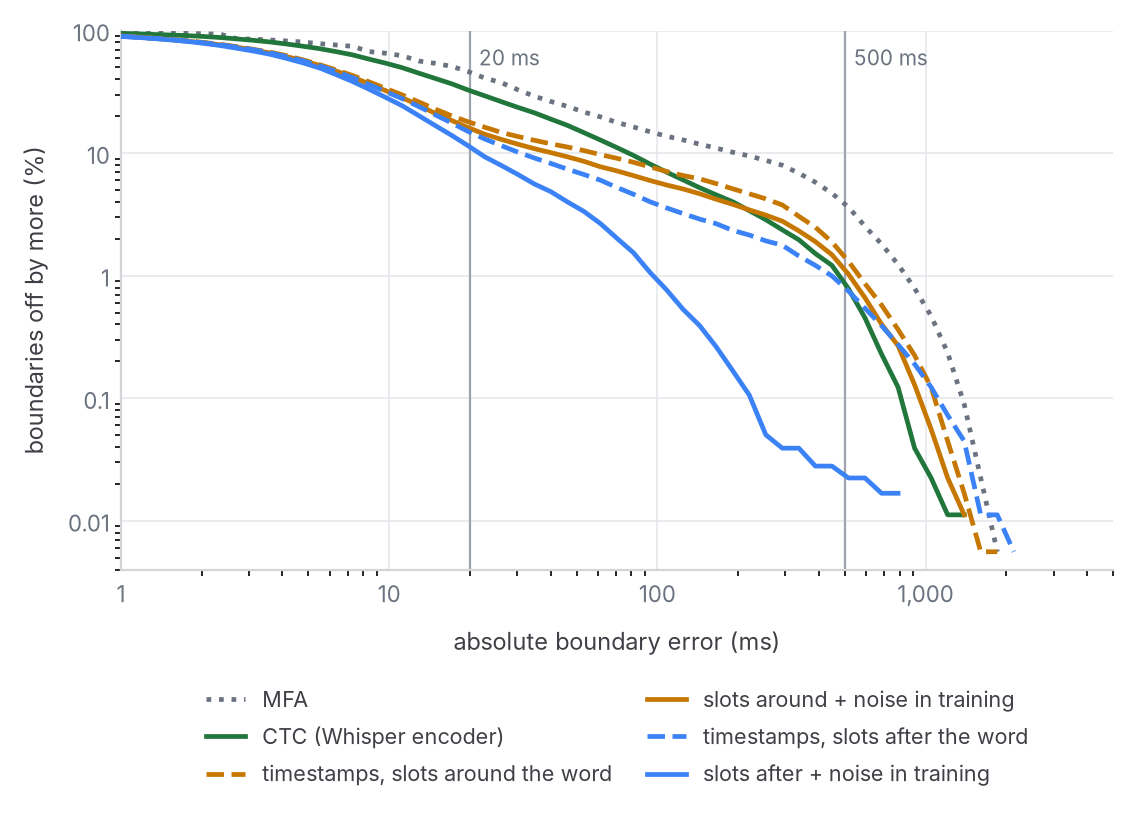}
\caption{Errors on the unchanged words of TIMIT at 20\% mixed transcript noise: the share of boundaries off by more
than each error, on log axes. Left of 20~ms, the timestamp models keep the most boundaries close. Right of 500~ms, the
model with the slots around the word has more gross errors than the CTC model. With the slots after the word and noise
in training, the gross errors are gone. The released large-v3 model's curve lies on that one.}
\label{fig:noisetail}
\end{figure}

\textbf{The timestamp model keeps the most boundaries within 20~ms under every kind of noise.} At 20\% mixed noise
on TIMIT it places \val{tn/timit/mix20/ts/w20}\% of them there, against \val{tn/timit/mix20/ctcw/w20}\% for the better
CTC model and \val{tn/timit/mix20/mfa/w20}\% for MFA, and its median error stays at
\val{tn/timit/mix20/ts/median}~ms. Figure~\ref{fig:noisetail} shows the whole distribution of the errors.

\begin{figure}[H]
\centering
\includegraphics[width=0.92\linewidth]{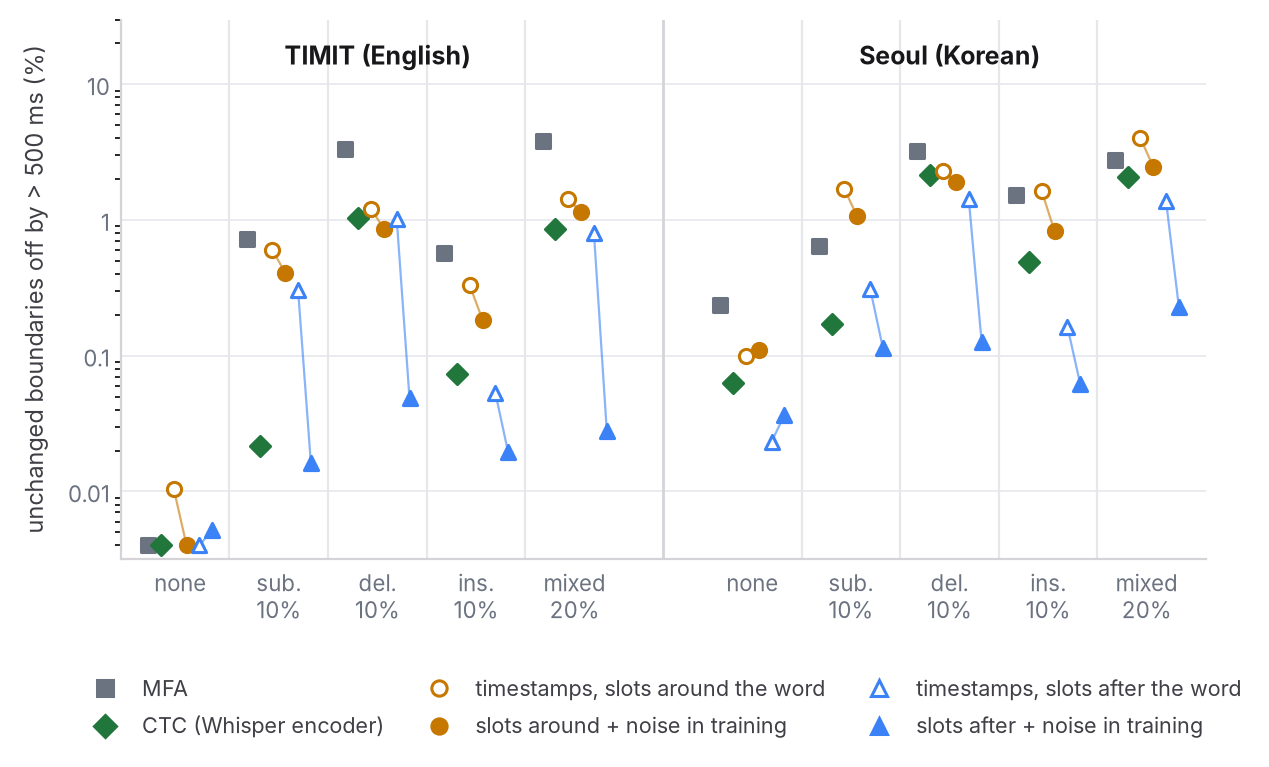}
\caption{Unchanged boundaries off by more than 500~ms, for each kind of transcript noise. \emph{None} is the clean
transcript, on the words the 10\% mixed copy leaves unchanged. Hollow markers are runs trained on clean transcripts and
filled ones runs trained with 10\% noise; a thin line joins the two runs of each slot layout. A point on the bottom edge
has no boundary that far off.}
\label{fig:noisekinds}
\end{figure}

\textbf{But its errors grow a heavy tail.} A substituted or inserted word sometimes throws neighbouring boundaries
more than half a second off: \val{tn/timit/sub10/ts/o500}\% of the unchanged boundaries on TIMIT with 10\%
substitutions, against \val{tn/timit/sub10/ctcw/o500}\% for the CTC model, and \val{tn/seoul/sub10/ts/o500}\% against
\val{tn/seoul/sub10/ctcw/o500}\% on Seoul (Figure~\ref{fig:noisekinds}). There the timestamp model's mean error
rises from \val{tn/seoul/sub10/ts/clean-mae} to \val{tn/seoul/sub10/ts/mae}~ms, while the CTC model's moves from
\val{tn/seoul/sub10/ctcw/clean-mae} to \val{tn/seoul/sub10/ctcw/mae}. A CTC path smears a wrong word over its own span;
the timestamp decoder reads the text and can lose its place. Deleted words hurt every aligner most, since the
neighbours stretch over the missing word, and MFA fails on whole utterances when words are inserted
(\val{tn/timit/ins10/mfa/failed} on TIMIT, \val{tn/seoul/ins10/mfa/failed} on Seoul).

\textbf{Training on noisy transcripts shrinks the tail.} We perturbed the training transcripts the same way: a
substituted word keeps the original word's times, a deleted word drops out, and an inserted word gets a zero-length
target in the middle of the gap after the word before it. With 10\% of words perturbed, unchanged boundaries off by
more than half a second on Seoul with substitutions fall from \val{tn/seoul/sub10/ts/o500}\% to
\val{tn/seoul/sub10/tn10/o500}\%, and at 20\% mixed noise the mean error falls from \val{tn/seoul/mix20/ts/mae} to
\val{tn/seoul/mix20/tn10/mae}~ms, level with the CTC model's \val{tn/seoul/mix20/ctcw/mae}, with
\val{tn/seoul/mix20/tn10/w20}\% of boundaries within 20~ms against its \val{tn/seoul/mix20/ctcw/w20}\%. On clean
transcripts it costs nothing: the seven-set mean moves by less than the seed spread. With the slots around the word,
the tail shrinks by a third to a half but does not vanish: with substitutions on Seoul the CTC model still has fewer
gross errors (\val{tn/seoul/sub10/ctcw/o500}\%).

\textbf{With the slots after the word, noise training also helps on clean transcripts.} Trained with 10\% noise in
that layout, the model scores \val{fin/tn-s5/mean-mae}~ms over the seven test sets, against \val{fin/post-s5/mean-mae}
without it (\val{fin/tn-s5/runs} and \val{fin/post-s5/runs} seeds; a difference of \val{fin/c/tn-s5/d}~ms, standard
error \val{fin/c/tn-s5/se}). Most of the gain is on Switchboard's conversations: \val{fin/tn-s5/swb/mae} against
\val{fin/post-s5/swb/mae}~ms. It costs a little on the codecs no augmentation uses (\val{fin/c/tn-s5/unseen/d}~ms).

\textbf{And it removes the tail.} With the slots after the word and noise in training, the error on a noisy
transcript hardly moves from the clean one. At 20\% mixed noise the released large-v3 model scores
\val{tn/timit/mix20/rel/mae}~ms on TIMIT's unchanged words, against \val{tn/timit/mix20/rel/clean-mae} with the clean
transcript, and \val{tn/seoul/mix20/rel/mae} against \val{tn/seoul/mix20/rel/clean-mae}~ms on Seoul
(Table~\ref{tab:noise}). Only \val{tn/timit/mix20/rel/o500}\% of its TIMIT boundaries are off by more than half a
second, against \val{tn/timit/mix20/ctcw/o500}\% for the CTC model and \val{tn/timit/mix20/mfa/o500}\% for MFA. The
slot order does part of this on its own: without noise training, the slots after the word already bring the default
model's error at that noise level from \val{tn/timit/mix20/ts/mae} to \val{tn/timit/mix20/post/mae}~ms.

\subsection{Release recipe}
\label{sec:release}

The released models put three changes together: the slots after the word, the call-centre degrader in place of the
six conditions, and 10\% transcript noise in training. We trained the steps toward this recipe on four checkpoints: the
release of Whisper-Scicom every earlier section used, a newer release fine-tuned on the same private data, and the
public large-v3 and large-v3-turbo (Table~\ref{tab:final}).

\begin{table}[H]
\centering
\caption{The release recipe, every arm with the slots after the word. \emph{Augmentation}: our six telephone conditions, or the call-centre degrader, and with 10\% transcript noise in training. Mean absolute error in ms, means over seeds; \emph{mean}: the seven test sets in their own condition, with the share within 20~ms; \emph{codecs}: TIMIT through the six codecs the conditions use; \emph{unseen}: TIMIT through AMR-NB and G.723.1, which no augmentation uses.}
\label{tab:final}
\scriptsize
\setlength{\tabcolsep}{2.4pt}
\begin{adjustbox}{max width=\linewidth}
\begin{tabular}{llrrrrrrrrrr}
\toprule
\schead \scth{checkpoint} & \scth{augmentation} & \scth{seeds} & \scth{mean} & \scth{TIMIT} & \scth{codecs} & \scth{unseen} & \scth{Buckeye} & \scth{SWB} & \scth{Seoul} & \scth{IFA} & \scth{VoxAng.} \\
\midrule
Whisper-Scicom, earlier & six conditions & 5 & 13.02 / 85.0 & 10.5 & 11.2 & 12.2 & 11.3 & 18.4 & 10.4 & 14.8 & 15.3 \\
Whisper-Scicom, newer & six conditions & 2 & 13.05 / 85.0 & 10.6 & 11.2 & 12.2 & 11.4 & 18.6 & 10.4 & 14.6 & 15.3 \\
large-v3 & six conditions & 2 & 12.88 / 85.2 & 10.4 & 11.0 & 12.1 & 11.3 & 18.3 & 10.4 & 14.6 & 15.0 \\
large-v3-turbo & six conditions & 2 & 12.99 / 84.9 & 10.4 & 11.0 & 12.0 & 11.5 & 17.7 & 10.5 & 14.8 & 15.5 \\
\midrule
Whisper-Scicom, earlier & six conditions + noise & 2 & 12.68 / 85.2 & 10.7 & 11.3 & 12.6 & 11.1 & 16.4 & 10.4 & 14.6 & 15.2 \\
\midrule
Whisper-Scicom, newer & call-centre degrader & 2 & 13.02 / 84.8 & 10.8 & 11.8 & 12.8 & 11.3 & 17.7 & 10.4 & 14.6 & 15.6 \\
Whisper-Scicom, earlier & call-centre degrader & 2 & 12.93 / 84.9 & 10.6 & 11.7 & 12.6 & 11.3 & 17.8 & 10.4 & 14.7 & 15.2 \\
\midrule
Whisper-Scicom, newer & degrader + noise & 2 & 12.88 / 85.0 & 10.8 & 12.0 & 13.0 & 11.2 & 16.2 & 10.5 & 14.5 & 16.4 \\
large-v3 & degrader + noise & 2 & 12.75 / 85.2 & 10.6 & 11.8 & 12.8 & 11.2 & 15.9 & 10.4 & 14.5 & 16.1 \\
large-v3-turbo & degrader + noise & 2 & 13.02 / 84.6 & 10.9 & 11.8 & 12.8 & 11.3 & 16.3 & 10.7 & 14.6 & 16.5 \\
\bottomrule
\end{tabular}
\end{adjustbox}
\end{table}

\textbf{The checkpoint still barely matters.} With the slots after the word and the six conditions, the newer
Whisper-Scicom scores \val{fin/post-s6/mean-mae}~ms over the seven test sets, large-v3 \val{fin/post-v3/mean-mae} and
turbo \val{fin/post-turbo/mean-mae}, against \val{fin/post-s5/mean-mae} for the earlier release. None of the
differences reaches two standard errors.

\textbf{The degrader buys robustness to degraded audio, not a measurable gain on Switchboard.} On the newer
Whisper-Scicom, the call-centre degrader in place of the six conditions moves Switchboard from
\val{fin/post-s6/swb/mae} to \val{fin/ce-s6/swb/mae}~ms over two seeds, a difference of
\val{fin/c/ce-vs-codecs/swb/d}~ms with a standard error of \val{fin/c/ce-vs-codecs/swb/se}; on the earlier release
the difference is \val{fin/c/ce-vs-codecs-s5/swb/d}~ms (standard error \val{fin/c/ce-vs-codecs-s5/swb/se}). It costs on
TIMIT's simulated codecs, from \val{fin/post-s6/codecs/mae} to \val{fin/ce-s6/codecs/mae}~ms through the six and from
\val{fin/post-s6/unseen/mae} to \val{fin/ce-s6/unseen/mae}~ms through the two neither augmentation uses, although the
six conditions there are the very codecs the other models trained on. What it buys shows through the degrader itself
(Table~\ref{tab:ce}).

\textbf{Transcript noise is what helps on real calls.} Added to the degrader, noise training moves Switchboard by
\val{fin/c/tn/swb/d}~ms (standard error \val{fin/c/tn/swb/se}); with the six conditions it moved it by
\val{fin/c/tn-s5/swb/d}~ms (standard error \val{fin/c/tn-s5/swb/se}, Section~\ref{sec:noise}). The full recipe ends at
\val{fin/final-v3/swb/mae}~ms on large-v3, \val{fin/final-turbo/swb/mae} on turbo and \val{fin/final-s6/swb/mae} on the
newer Whisper-Scicom, two seeds each, against \val{fin/post-v3/swb/mae}, \val{fin/post-turbo/swb/mae} and
\val{fin/post-s6/swb/mae} with the slots after the word alone.

\textbf{Released models.} We release the first seed of the recipe on large-v3 and on large-v3-turbo, fixed before any
test score was seen; the variant on Whisper-Scicom stays private with its base. On FA-Bench's TIMIT core test they
score \val{fb/rel-v3/mae} and \val{fb/rel-turbo/mae}~ms, at an F1 of \val{fb/rel-v3/f1} and \val{fb/rel-turbo/f1}
within 20~ms. On Switchboard's real calls they score \val{swb/ft-ts-v3-final/tel/mae} and
\val{swb/ft-ts-turbo-final/tel/mae}~ms, and through the call-centre degrader \val{ce/rel-v3/timit/ce/mae} and
\val{ce/rel-turbo/timit/ce/mae}~ms on TIMIT (Section~\ref{sec:realcalls}).

\section{Conclusion}
\label{sec:conclusion}

\begin{figure}[H]
\centering
\includegraphics[width=0.9\linewidth]{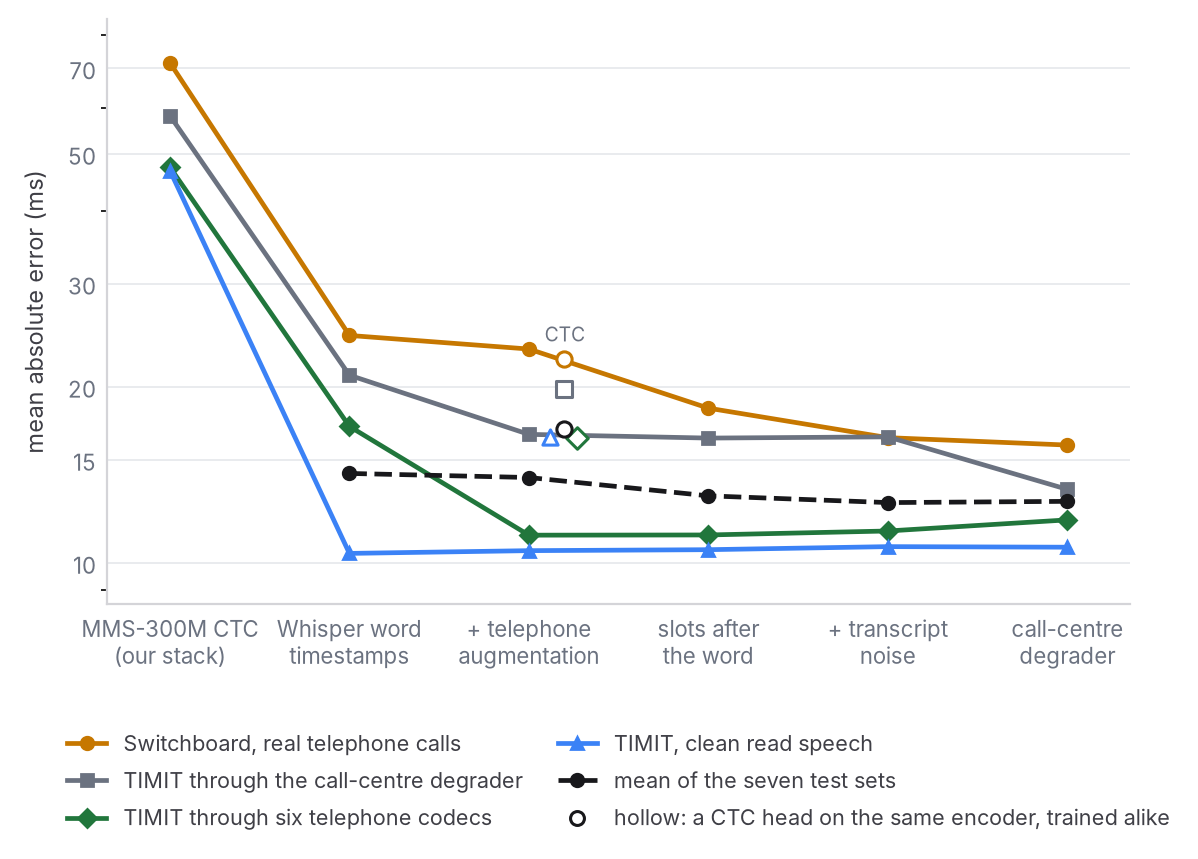}
\caption{How each change moved the error, on a log scale. Each step adds one change to the one before: our stack's
MMS-300M CTC aligner as it runs; Whisper predicting word timestamps, trained on human boundaries; telephone
augmentation; the slots after the word; transcript noise in training; the call-centre degrader in place of the six
channels. The last point is the released large-v3 model and the others are Whisper-Scicom, a difference that changes
nothing measurable (Section~\ref{sec:checkpoint}). Hollow markers: a CTC head on the same encoder, trained like the
third step. Means over each arm's seeds (\val{prog/ts/n}, \val{prog/aug/n}, \val{prog/post/n}, \val{prog/noise/n} and
\val{prog/degrader/n}).}
\label{fig:progress}
\end{figure}

We set out to replace the aligner in our own stack with one that is precise against human boundaries, works across
languages and holds up on telephone speech. Five changes moved the error, each measured against the one before
(Figure~\ref{fig:progress}).

\begin{enumerate}
\item \textbf{Word timestamps from an encoder-decoder instead of a CTC path.} Trained on human boundaries, Whisper
predicting each word's start and end as timestamp tokens scores \val{prog/ts/clean}~ms on clean TIMIT, where the
MMS-300M CTC aligner our stack runs scores \val{prog/stack/clean}~ms. The head is what matters: a CTC head on the same
encoder, trained the same way, scores \val{prog/ctc/seven}~ms over the seven test sets against the timestamp model's
\val{prog/aug/seven}.
\item \textbf{Telephone augmentation.} Passing half the training clips through six simulated telephone channels brings
TIMIT through those codecs from \val{prog/ts/codecs} to \val{prog/aug/codecs}~ms and costs nothing on clean speech.
\item \textbf{Timestamp slots after the word.} The decoder reads a word before it times it. Switchboard's telephone
conversations go from \val{prog/aug/swb} to \val{prog/post/swb}~ms, and the seven-set mean from \val{prog/aug/seven} to
\val{prog/post/seven}.
\item \textbf{Transcript noise in training.} With 10\% of the training words substituted, deleted or inserted,
Switchboard falls to \val{prog/noise/swb}~ms and the seven-set mean to \val{prog/noise/seven}, and a wrong word in the
transcript no longer throws its neighbours half a second off (Section~\ref{sec:noise}).
\item \textbf{The call-centre degrader.} Trained on audio degraded like real calls in place of the six channels, the
released large-v3 model scores \val{prog/degrader/ce}~ms on TIMIT through that degrader, against \val{prog/noise/ce},
and \val{prog/degrader/swb}~ms on Switchboard.
\end{enumerate}

Two choices hold up every step: training only on boundaries people placed, and training on many languages, which cuts
the error on the held-out languages by \val{tr/cut-min} to \val{tr/cut-max}\% against English alone
(Section~\ref{sec:transfer}). Much else did not move the error: the adapters' rank, learning rate and training length,
full fine-tuning, a boundary head on the encoder, the Whisper checkpoint (Section~\ref{sec:checkpoint}) and finer
timestamp tokens (Section~\ref{sec:finer}).

\section{Limitations}

\textbf{English ground truth in the pilot.} All three human-labelled sets in the pilot are English, so whatever the
pilot says about other languages is inference.

\textbf{No external test.} The trained models are tested on held-out speakers and recordings of the corpora they
trained on, in English, Korean, Dutch and the VoxAngeles languages, so they know each test set's labelling convention
and MFA does not. Korean has \val{tr/matched/seoul/speakers} test speakers, Dutch \val{tr/matched/ifa/speakers}, and VoxAngeles one per language, with one word per clip.

\textbf{Mostly simulated telephone channels.} Apart from Switchboard, the codecs are real and the line is not. Real
networks add packet loss, transcoding chains, echo cancellation and automatic gain control, which the six simulated
conditions do not model.

\textbf{Defaults only in the pilot.} Every existing aligner ran at its documented defaults. Several can be tuned, MFA
most of all, and a tuned configuration may rank differently.

\textbf{Few seeds.} The adapter sweep reads each factor against \val{s3/n-seeds} seeds of the default, and runs every
other configuration once. Telephone augmentation cannot be repeated exactly: its random draws are not seeded.

\textbf{Mirrors and licences.} TIMIT (LDC93S1) and Buckeye are read from third-party Hugging Face copies, and the
Switchboard excerpts from a public mirror of LDC-licensed audio. Results are for internal study until the licence
position is confirmed.

\section{Acknowledgement}

This work was carried out at Scicom (MSC) Berhad. The author thanks Scicom for the compute and for the call-centre
work that made the problem concrete.

\bibliographystyle{plainnat}
\bibliography{neurips_2023}

\clearpage
\appendix
\section{Transfer from English, every language}
\label{app:transferlang}

{\scriptsize
\setlength{\tabcolsep}{3.2pt}
\begin{longtable}{llrrccrrrrr}
\caption{Transfer from English, every language: the timestamp recipe trained on the English rows only (\emph{English}) and on all training languages (\emph{all}), on the same utterances. $\Delta$: the change in the share of boundaries within 20~ms. Korean and Dutch are held-out speakers of the Seoul Corpus and IFA; the other rows are VoxAngeles, one speaker per language and one word per clip, so a row rests on a handful of boundaries (\emph{bound.}: two per clip, so a language with one clip can only score 0, 50 or 100\%). \emph{Trained}: the language is in the multilingual training (other speakers); \emph{Whisper}: Whisper's pre-training covers it.}
\label{tab:transferlang}\\
\toprule
\schead & & & & & & \multicolumn{3}{c}{\scth{within 20~ms (\%)}} & \multicolumn{2}{c}{\scth{MAE (ms)}} \\
\schead \scth{language} & \scth{code} & \scth{clips} & \scth{bound.} & \scth{trained} & \scth{Whisper} & \scth{English} & \scth{all} & \scth{$\Delta$} & \scth{English} & \scth{all} \\
\midrule
\endfirsthead
\toprule
\schead & & & & & & \multicolumn{3}{c}{\scth{within 20~ms (\%)}} & \multicolumn{2}{c}{\scth{MAE (ms)}} \\
\schead \scth{language} & \scth{code} & \scth{clips} & \scth{bound.} & \scth{trained} & \scth{Whisper} & \scth{English} & \scth{all} & \scth{$\Delta$} & \scth{English} & \scth{all} \\
\midrule
\endhead
\bottomrule
\endlastfoot
Korean (Seoul Corpus) & kor & 1{,}294 & 44{,}540 & yes & yes & 79 & 90 & $+11$ & 24.2 & 11.3 \\
Dutch (IFA) & nld & 333 & 5{,}028 & yes & yes & 72 & 78 & $+7$ & 22.1 & 15.5 \\
\midrule
Abkhazian & abk & 5 & 10 & yes & no & 100 & 100 & 0 & 8.3 & 6.1 \\
Achinese & ace & 8 & 16 & yes & no & 50 & 75 & $+25$ & 49.9 & 22.3 \\
Adamawa Fulfulde & fub & 1 & 2 & yes & no & 0 & 0 & 0 & 35.4 & 37.2 \\
Adyghe & ady & 116 & 232 & no & no & 50 & 73 & $+23$ & 44.1 & 22.5 \\
Afrikaans & afr & 15 & 30 & yes & yes & 57 & 80 & $+23$ & 24.0 & 14.1 \\
Aghul & agx & 4 & 8 & yes & no & 38 & 38 & 0 & 31.3 & 22.5 \\
Akan & aka & 4 & 8 & yes & no & 75 & 100 & $+25$ & 15.9 & 6.3 \\
Albay Bikol & bhk & 3 & 6 & yes & no & 33 & 50 & $+17$ & 57.9 & 19.5 \\
Angami Naga & njm & 7 & 14 & yes & no & 43 & 43 & 0 & 37.3 & 40.1 \\
Armenian & hye & 7 & 14 & yes & yes & 64 & 93 & $+29$ & 20.7 & 8.4 \\
Assamese & asm & 11 & 22 & yes & yes & 68 & 91 & $+23$ & 20.7 & 11.8 \\
Bafut & bfd & 2 & 4 & yes & no & 50 & 100 & $+50$ & 54.6 & 10.4 \\
Bambara & bam & 8 & 16 & yes & no & 75 & 81 & $+6$ & 15.4 & 12.4 \\
Basque & eus & 42 & 84 & no & yes & 85 & 96 & $+12$ & 12.0 & 7.2 \\
Bassa & bsq & 2 & 4 & yes & no & 25 & 50 & $+25$ & 91.1 & 18.5 \\
Bemba (Zambia) & bem & 4 & 8 & yes & no & 88 & 100 & $+12$ & 9.9 & 5.7 \\
Bengali & ben & 1 & 2 & yes & yes & 100 & 100 & 0 & 7.5 & 6.6 \\
Bini & bin & 6 & 12 & yes & no & 50 & 50 & 0 & 30.8 & 26.0 \\
Bukiyip & ape & 8 & 16 & yes & no & 69 & 75 & $+6$ & 28.0 & 24.1 \\
Bura-Pabir & bwr & 2 & 4 & yes & no & 75 & 75 & 0 & 24.1 & 14.5 \\
Burmese & mya & 12 & 24 & yes & yes & 58 & 83 & $+25$ & 29.6 & 19.1 \\
Cacua & cbv & 5 & 10 & yes & no & 50 & 70 & $+20$ & 40.2 & 17.9 \\
Central Atlas Tamazight & tzm & 2 & 4 & yes & no & 50 & 75 & $+25$ & 115.9 & 16.4 \\
Chamalal & cji & 5 & 10 & yes & no & 70 & 80 & $+10$ & 17.3 & 32.4 \\
Chamorro & cha & 5 & 10 & yes & no & 70 & 100 & $+30$ & 24.2 & 6.1 \\
Cherepon & cpn & 11 & 22 & yes & no & 55 & 73 & $+18$ & 27.0 & 16.5 \\
Columbia-Wenatchi & col & 6 & 12 & yes & no & 50 & 67 & $+17$ & 25.4 & 25.4 \\
Croatian & hrv & 4 & 8 & yes & yes & 88 & 100 & $+12$ & 8.3 & 6.9 \\
Czech & ces & 4 & 8 & yes & yes & 75 & 88 & $+12$ & 17.6 & 10.2 \\
Dagbani & dag & 21 & 42 & no & no & 48 & 90 & $+43$ & 57.2 & 10.2 \\
Danish & dan & 8 & 16 & yes & yes & 38 & 94 & $+56$ & 42.6 & 8.2 \\
Dari & prs & 9 & 18 & yes & yes & 78 & 83 & $+6$ & 14.3 & 10.7 \\
Defaka & afn & 5 & 10 & yes & no & 70 & 100 & $+30$ & 12.0 & 7.3 \\
Degema & deg & 4 & 8 & yes & no & 75 & 100 & $+25$ & 11.3 & 5.6 \\
Dutch & nld & 10 & 20 & yes & yes & 75 & 80 & $+5$ & 15.1 & 12.9 \\
Efik & efi & 6 & 12 & yes & no & 58 & 83 & $+25$ & 38.9 & 13.9 \\
Emai-Iuleha-Ora & ema & 3 & 6 & yes & no & 50 & 67 & $+17$ & 77.4 & 17.4 \\
Ewe & ewe & 5 & 10 & yes & no & 30 & 70 & $+40$ & 52.2 & 12.7 \\
Finnish & fin & 8 & 16 & yes & yes & 62 & 94 & $+31$ & 22.2 & 8.7 \\
Ga & gaa & 3 & 6 & yes & no & 67 & 83 & $+17$ & 32.0 & 13.4 \\
Ganda & lug & 5 & 10 & yes & no & 50 & 60 & $+10$ & 23.2 & 20.0 \\
Gua & gwx & 1 & 2 & yes & no & 0 & 50 & $+50$ & 110.6 & 13.0 \\
Gujarati & guj & 7 & 14 & yes & yes & 57 & 93 & $+36$ & 13.8 & 8.7 \\
Hakka Chinese & hak & 5 & 10 & yes & no & 60 & 90 & $+30$ & 13.8 & 8.3 \\
Hausa & hau & 58 & 116 & no & yes & 72 & 86 & $+14$ & 18.2 & 11.1 \\
Hawaiian & haw & 8 & 16 & yes & yes & 62 & 62 & 0 & 26.3 & 20.4 \\
Hebrew & heb & 3 & 6 & yes & yes & 100 & 100 & 0 & 7.7 & 6.4 \\
Hiligaynon & hil & 5 & 10 & yes & no & 50 & 70 & $+20$ & 33.4 & 24.6 \\
Hindi & hin & 7 & 14 & yes & yes & 64 & 86 & $+21$ & 16.1 & 10.6 \\
Hungarian & hun & 10 & 20 & yes & yes & 85 & 85 & 0 & 12.4 & 8.4 \\
Ibibio & ibb & 60 & 120 & no & no & 61 & 78 & $+18$ & 28.1 & 17.5 \\
Icelandic & isl & 11 & 22 & yes & yes & 64 & 86 & $+23$ & 28.7 & 9.5 \\
Idoma & idu & 4 & 8 & yes & no & 75 & 100 & $+25$ & 12.1 & 4.6 \\
Igbo & ibo & 3 & 6 & yes & no & 67 & 83 & $+17$ & 17.4 & 11.8 \\
Iloko & ilo & 7 & 14 & yes & no & 86 & 93 & $+7$ & 10.6 & 6.4 \\
Iranian Persian & pes & 14 & 28 & yes & yes & 75 & 89 & $+14$ & 15.9 & 10.2 \\
Jola-Fonyi & dyo & 4 & 8 & yes & no & 38 & 88 & $+50$ & 33.3 & 15.2 \\
Kabuverdianu & kea & 3 & 6 & yes & no & 67 & 83 & $+17$ & 18.5 & 9.4 \\
Kannada & kan & 10 & 20 & yes & yes & 85 & 100 & $+15$ & 9.9 & 6.6 \\
Kele (Papua New Guinea) & sbc & 4 & 8 & yes & no & 38 & 88 & $+50$ & 25.9 & 14.5 \\
Khmer & khm & 60 & 120 & no & yes & 64 & 87 & $+22$ & 26.3 & 17.9 \\
Klao & klu & 6 & 12 & yes & no & 50 & 58 & $+8$ & 22.1 & 18.1 \\
Konkani (individual language) & knn & 6 & 12 & yes & no & 67 & 83 & $+17$ & 22.5 & 7.7 \\
Koonzime & ozm & 2 & 4 & yes & no & 25 & 25 & 0 & 49.5 & 45.6 \\
Krache & kye & 2 & 4 & yes & no & 25 & 100 & $+75$ & 41.1 & 6.7 \\
Krio & kri & 1 & 2 & yes & no & 0 & 100 & $+100$ & 50.0 & 3.5 \\
Ladino & lad & 55 & 110 & no & no & 46 & 75 & $+29$ & 36.3 & 13.6 \\
Lakota & lkt & 2 & 4 & yes & no & 25 & 75 & $+50$ & 67.4 & 12.7 \\
Larteh & lar & 1 & 2 & yes & no & 100 & 100 & 0 & 5.3 & 6.4 \\
Latvian & lav & 7 & 14 & yes & yes & 64 & 79 & $+14$ & 21.9 & 20.7 \\
Lendu & led & 1 & 2 & yes & no & 50 & 50 & 0 & 39.4 & 30.6 \\
Levantine Arabic & apc & 12 & 24 & yes & yes & 62 & 67 & $+4$ & 27.1 & 27.6 \\
Lithuanian & lit & 18 & 36 & yes & yes & 67 & 89 & $+22$ & 17.2 & 10.9 \\
Logba & lgq & 24 & 48 & no & no & 56 & 77 & $+21$ & 27.5 & 13.0 \\
Maasina Fulfulde & ffm & 1 & 2 & yes & no & 100 & 100 & 0 & 5.3 & 7.8 \\
Makasar & mak & 5 & 10 & yes & no & 60 & 70 & $+10$ & 44.4 & 25.1 \\
Malayalam & mal & 2 & 4 & yes & yes & 75 & 100 & $+25$ & 11.6 & 6.0 \\
Maltese & mlt & 93 & 186 & no & yes & 58 & 66 & $+8$ & 28.2 & 21.0 \\
Min Nan Chinese & nan & 74 & 148 & no & no & 71 & 84 & $+13$ & 24.9 & 19.3 \\
Modern Greek (1453-) & ell & 7 & 14 & yes & yes & 100 & 100 & 0 & 5.9 & 4.5 \\
Pampanga & pam & 2 & 4 & yes & no & 75 & 75 & 0 & 9.1 & 16.3 \\
Rundi & run & 6 & 12 & yes & no & 42 & 83 & $+42$ & 27.8 & 12.6 \\
Scottish Gaelic & gla & 6 & 12 & yes & no & 67 & 92 & $+25$ & 43.7 & 11.0 \\
South Azerbaijani & azb & 3 & 6 & yes & yes & 50 & 67 & $+17$ & 45.7 & 44.1 \\
South Levantine Arabic & ajp & 6 & 12 & yes & yes & 58 & 92 & $+33$ & 40.3 & 8.6 \\
Tsishingini & tsw & 2 & 4 & yes & no & 100 & 100 & 0 & 4.6 & 9.1 \\
Tunisian Arabic & aeb & 5 & 10 & yes & yes & 50 & 50 & 0 & 36.0 & 34.3 \\
Western Apache & apw & 6 & 12 & yes & no & 67 & 92 & $+25$ & 17.4 & 7.9 \\
Wu Chinese & wuu & 3 & 6 & yes & no & 67 & 100 & $+33$ & 15.2 & 7.7 \\
Yue Chinese & yue & 16 & 32 & yes & yes & 59 & 91 & $+31$ & 33.9 & 9.4 \\
\end{longtable}
}

\end{document}